\documentclass{article}

\PassOptionsToPackage{round, sort}{natbib}

\usepackage[final]{neurips_2021}

\usepackage[utf8]{inputenc} 
\usepackage[T1]{fontenc}    
\usepackage{hyperref}       
\usepackage{url}            
\usepackage{booktabs}       
\usepackage{amsfonts}       
\usepackage{nicefrac}       
\usepackage{microtype}      
\usepackage{xcolor}         

\usepackage[ruled]{algorithm2e}
\usepackage{amsmath}
\usepackage{colortbl}
\usepackage{enumitem}
\usepackage{float}
\usepackage{graphicx}
\usepackage{hypcap}
\usepackage{listings}
\usepackage{mathtext}
\usepackage{pgfplots}  
\usepackage{rotating,tabularx}
\usepackage{siunitx}
\usepackage{subcaption}
\usepackage{wrapfig}  

\graphicspath{ {./data/} }

\hypersetup{
    colorlinks,
    allcolors=.,
    urlcolor=blue,
}
\pgfplotsset{compat=newest}
\usepgfplotslibrary{groupplots,dateplot}
\usetikzlibrary{patterns,shapes.arrows}
\usepackage{lib}

\title{Revisiting Deep Learning Models for Tabular Data}

\author{%
    Yury Gorishniy\thanks{The first author: \texttt{firstnamelastname@gmail.com}} \textsuperscript{\hspace{0.3em}$\dag \ddag$}
    \And
    Ivan Rubachev\textsuperscript{$\dag \clubsuit$}
    \And
    Valentin Khrulkov\textsuperscript{$\dag$}
    \And    
    Artem Babenko\textsuperscript{$\dag \clubsuit$}
    \and\\
    \centerline{$\dag$ Yandex}\\
    \centerline{$\ddag$ Moscow Institute of Physics and Technology}\\
    \centerline{$\clubsuit$ National Research University Higher School of Economics}
\vspace{-16pt}
}

\begin{document}

\maketitle

\setcounter{footnote}{0}

\begin{abstract}
    The existing literature on deep learning for tabular data proposes a wide range of novel architectures and reports competitive results on various datasets.
    However, the proposed models are usually not properly compared to each other and existing works often use different benchmarks and experiment protocols.
    As a result, it is unclear for both researchers and practitioners what models perform best.
    Additionally, the field still lacks effective baselines, that is, the easy-to-use models that provide competitive performance across different problems.

    In this work, we perform an overview of the main families of DL architectures for tabular data and raise the bar of baselines in tabular DL by identifying two simple and powerful deep architectures.
    The first one is a ResNet-like architecture which turns out to be a strong baseline that is often missing in prior works.
    The second model is our simple adaptation of the Transformer architecture for tabular data, which outperforms other solutions on most tasks.
    Both models are compared to many existing architectures on a diverse set of tasks under the same training and tuning protocols.
    We also compare the best DL models with Gradient Boosted Decision Trees and conclude that there is still no universally superior solution.
    The source code is available at \url{\repository}.
\end{abstract}

\section{Introduction}

Due to the tremendous success of deep learning on such data domains as images, audio and texts \citep{goodfellow-book}, there has been a lot of research interest to extend this success to problems with data stored in tabular format.
In these problems, data points are represented as vectors of heterogeneous features, which is typical for industrial applications and ML competitions, where neural networks have a strong non-deep competitor in the form of GBDT \citep{xgboost, catboost, lightgbm}.
Along with potentially higher performance, using deep learning for tabular data is appealing as it would allow constructing multi-modal pipelines for problems, where only one part of the input is tabular, and other parts include images, audio and other DL-friendly data.
Such pipelines can then be trained end-to-end by gradient optimization for all modalities.
For these reasons, a large number of DL solutions were recently proposed, and new models continue to emerge \citep{snn, node, tabnet, autoint, dcn, dcn2, grownet, tel, tabtransformer}.

Unfortunately, due to the lack of established benchmarks (such as ImageNet \citep{imagenet} for computer vision or GLUE \citep{glue} for NLP), existing papers use different datasets for evaluation and proposed DL models are often not adequately compared to each other.
Therefore, from the current literature, it is unclear what DL model generally performs better than others and whether GBDT is surpassed by DL models.
Additionally, despite the large number of novel architectures, the field still lacks simple and reliable solutions that allow achieving competitive performance with moderate effort and provide stable performance across many tasks.
In that regard, Multilayer Perceptron (MLP) remains the main simple baseline for the field, however, it does not always represent a significant challenge for other competitors.

The described problems impede the research process and make the observations from the papers not conclusive enough.
Therefore, we believe it is timely to review the recent developments from the field and raise the bar of baselines in tabular DL.
We start with a hypothesis that well-studied DL architecture blocks may be underexplored in the context of tabular data and may be used to design better baselines.
Thus, we take inspiration from well-known battle-tested architectures from other fields and obtain two simple models for tabular data.
The first one is a ResNet-like architecture \citep{resnet} and the second one is FT-Transformer --- our simple adaptation of the Transformer architecture \citep{transformer} for tabular data.
Then, we compare these models with many existing solutions on a diverse set of tasks under the same protocols of training and hyperparameters tuning.
First, we reveal that none of the considered DL models can consistently outperform the ResNet-like model.
Given its simplicity, it can serve as a strong baseline for future work.
Second, FT-Transformer demonstrates the best performance on most tasks and becomes a new powerful solution for the field.
Interestingly, FT-Transformer turns out to be a more universal architecture for tabular data: it performs well on a wider range of tasks than the more ``conventional'' ResNet and other DL models.
Finally, we compare the best DL models to GBDT and conclude that there is still no universally superior solution.

We summarize the contributions of our paper as follows:
\begin{enumerate}
    \item We thoroughly evaluate the main models for tabular DL on a diverse set of tasks to investigate their relative performance.
    \item We demonstrate that a simple ResNet-like architecture is an effective baseline for tabular DL, which was overlooked by existing literature. Given its simplicity, we recommend this baseline for comparison in future tabular DL works.
    \item We introduce FT-Transformer --- a simple adaptation of the Transformer architecture for tabular data that becomes a new powerful solution for the field. We observe that it is a more universal architecture: it performs well on a wider range of tasks than other DL models.
    \item We reveal that there is still no universally superior solution among GBDT and deep models.
\end{enumerate}

\section{Related work}

\textbf{The ``shallow'' state-of-the-art} for problems with tabular data is currently ensembles of decision trees, such as GBDT (Gradient Boosting Decision Tree) \citep{greedy-func-approx}, which are typically the top-choice in various ML competitions.
At the moment, there are several established GBDT libraries, such as XGBoost  \citep{xgboost}, LightGBM  \citep{lightgbm}, CatBoost  \citep{catboost}, which are widely used by both ML researchers and practitioners.
While these implementations vary in detail, on most of the tasks, their performances do not differ much  \citep{catboost}. 

During several recent years, a large number of deep learning models for tabular data have been developed \citep{snn, node, tabnet, autoint, dcn, grownet, tel, tabtransformer}.
Most of these models can be roughly categorized into three groups, which we briefly describe below.

\textbf{Differentiable trees.}
The first group of models is motivated by the strong performance of decision tree ensembles for tabular data.
Since decision trees are not differentiable and do not allow gradient optimization, they cannot be used as a component for pipelines trained in the end-to-end fashion.
To address this issue, several works \citep{dndf, dndt,node,tel} propose to ``smooth'' decision functions in the internal tree nodes to make the overall tree function and tree routing differentiable.
While the methods of this family can outperform GBDT on some tasks \citep{node}, in our experiments, they do not consistently outperform ResNet.

\textbf{Attention-based models.}
Due to the ubiquitous success of attention-based architectures for different domains \citep{transformer, vit}, several authors propose to employ attention-like modules for tabular DL as well \citep{tabnet, autoint, tabtransformer}.
In our experiments, we show that the properly tuned ResNet outperforms the existing attention-based models.
Nevertheless, we identify an effective way to apply the Transformer architecture \citep{transformer} to tabular data: the resulting architecture outperforms ResNet on most of the tasks.

\textbf{Explicit modeling of multiplicative interactions.}
In the literature on recommender systems and click-through-rate prediction, several works criticize MLP since it is unsuitable for modeling multiplicative interactions between features \citep{latent-cross,dcn,dasalc}.
Inspired by this motivation, some works \citep{latent-cross,dcn,dcn2} have proposed different ways to incorporate feature products into MLP.
In our experiments, however, we do not find such methods to be superior to properly tuned baselines.

The literature also proposes some other architectural designs \citep{grownet, snn} that cannot be explicitly assigned to any of the groups above.
Overall, the community has developed a variety of models that are evaluated on different benchmarks and are rarely compared to each other.
Our work aims to establish a fair comparison of them and identify the solutions that consistently provide high performance.

\vfill

\section{Models for tabular data problems}
In this section, we describe the main deep architectures that we highlight in our work, as well as the existing solutions included in the comparison.
Since we argue that the field needs strong easy-to-use baselines, we try to reuse well-established DL building blocks as much as possible when designing ResNet (section 3.2) and FT-Transformer (section 3.3).
We hope this approach will result in conceptually familiar models that require less effort to achieve good performance.
Additional discussion and technical details for all the models are provided in supplementary.

\textbf{Notation.}
In this work, we consider supervised learning problems.
$D{=}\{(x_i,\ y_i)\}_{i = 1}^{n}$ denotes a dataset, where $x_i{=}(x_i^{(num)},\ x_i^{(cat)}) \in \X$ represents numerical $x_{ij}^{(num)}$ and categorical $x_{ij}^{(cat)}$ features of an object and $y_i \in \Y$ denotes the corresponding object label.
The total number of features is denoted as $k$.
The dataset is split into three disjoint subsets: $D = D_{train} \ \cup \ D_{val}\ \cup \ D_{test}$, where $D_{train}$ is used for training, $D_{val}$ is used for early stopping and hyperparameter tuning, and $D_{test}$ is used for the final evaluation.
We consider three types of tasks: binary classification $\Y = \{0,\ 1\}$, multiclass classification $\Y = \{1,\ \ldots,\ C\}$ and regression $\Y = \R$.

\vfill

\subsection{MLP}
We formalize the ``MLP'' architecture in \autoref{eq:mlp}.

\vfill

\begin{equation} \label{eq:mlp}
\begin{aligned}
    \texttt{MLP}(x) &= \Linear \left(
        \texttt{MLPBlock} \left( \ldots \left( \texttt{MLPBlock}(x) \right) \right)
    \right) \\
    \texttt{MLPBlock}(x) &= \Dropout(\ReLU(\Linear(x)))
\end{aligned}
\end{equation}

\vfill

\subsection{ResNet}
\label{sec:resnet}
We are aware of one attempt to design a ResNet-like baseline \citep{snn} where the reported results were not competitive. However, given ResNet's success story in computer vision \citep{resnet} and its recent achievements on NLP tasks \citep{resnet-nlp}, we give it a second try and construct a simple variation of ResNet as described in \autoref{eq:resnet}. The main building block is simplified compared to the original architecture, and there is an almost clear path from the input to output which we find to be beneficial for the optimization. Overall, we expect this architecture to outperform MLP on tasks where deeper representations can be helpful.

\vfill

\begin{equation} \label{eq:resnet}
\begin{aligned}
    \texttt{ResNet}(x) &= \texttt{Prediction} \left(
        \texttt{ResNetBlock} \left( \ldots \left( \texttt{ResNetBlock} \left( \Linear (x) \right) \right) \right)
    \right) \\
    \texttt{ResNetBlock}(x) &= x + \Dropout(\Linear(\Dropout(\ReLU(\Linear(\BatchNorm(x)))))) \\
    \texttt{Prediction}(x) &= \Linear \left( \ReLU \left( \BatchNorm \left( x \right) \right) \right)
\end{aligned}
\end{equation}

\vfill

\subsection{\architecture}
\label{sec:transformer}

\begin{figure}[ht]
    \centering
    \includegraphics[width=0.9\linewidth]{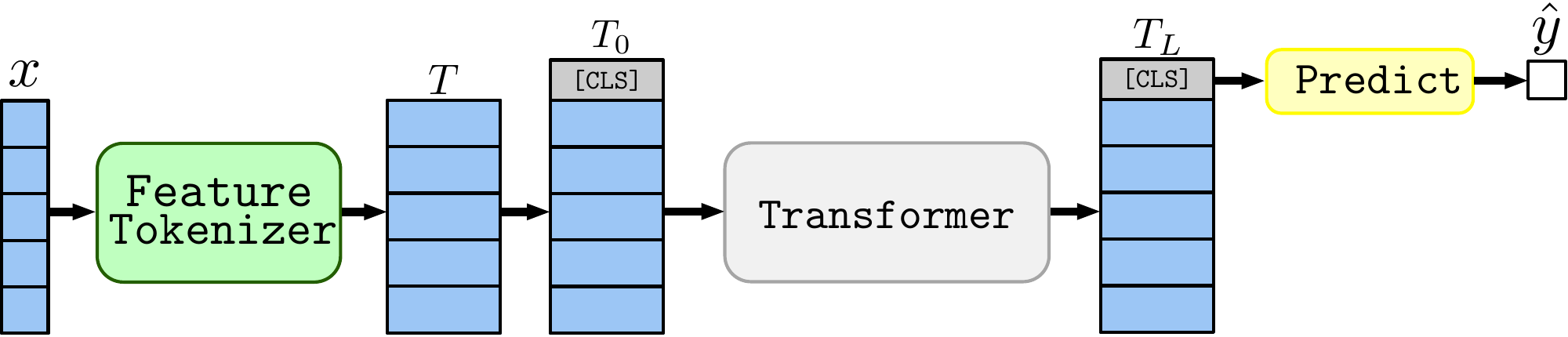}
    \caption{The \architecture\ architecture. Firstly, \tokenizer\ transforms features to embeddings. The embeddings are then processed by the Transformer module and the final representation of the \CLS\ token is used for prediction.}
    \label{fig:arch}
\end{figure}

In this section, we introduce \architecture\ (\textbf{F}eature \textbf{T}okenizer + \textbf{Transformer}) --- a simple adaptation of the Transformer architecture \citep{transformer} for the tabular domain. \autoref{fig:arch} demonstrates the main parts of \architecture.
In a nutshell, our model transforms all features (categorical and numerical) to embeddings and applies a stack of Transformer layers to the embeddings.
Thus, every Transformer layer operates on the \textit{feature} level of \textit{one} object.
We compare \architecture\ to conceptually similar AutoInt in \autoref{sec:ablation}.

\begin{figure}[ht]
    \centering
    \includegraphics[width=0.9\linewidth]{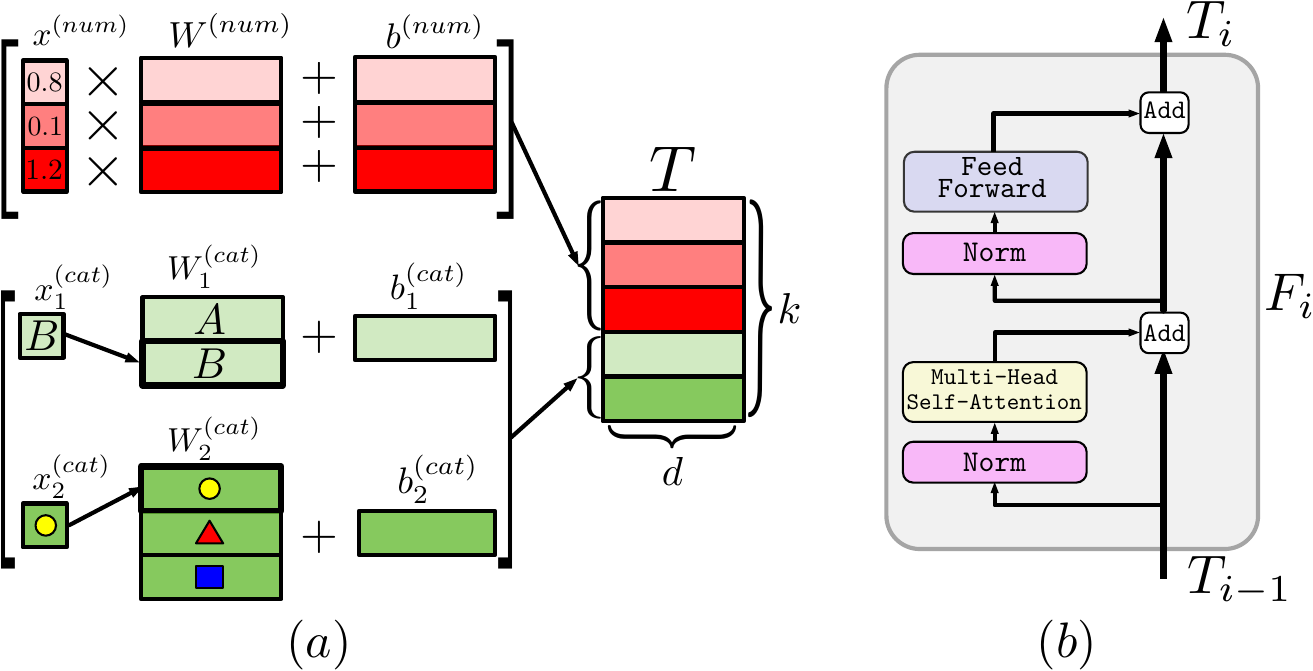}
    \caption{(a) \tokenizer; in the example, there are three numerical and two categorical features; (b) One Transformer layer.}
    \label{fig:blocks}
\end{figure}

\textbf{\tokenizer.} The \tokenizer\ module (see \autoref{fig:blocks}) transforms the input features $x$ to embeddings $T \in \R^{k \times d}$. The embedding for a given feature $x_j$ is computed as follows:
$$
    T_j = b_j + f_j(x_j) \in \R^d \qquad f_j: \X_j \rightarrow \R^d.
$$
where $b_j$ is the $j$-th \textit{feature bias}, $f^{(num)}_j$ is implemented as the element-wise multiplication with the vector \mbox{$W^{(num)}_j \in \R^d$} and $f^{(cat)}_j$ is implemented as the lookup table \mbox{$W^{(cat)}_j \in \R^{S_{j} \times d}$} for categorical features. Overall:
\begin{alignat*}{2}
    &T^{(num)}_j  = b^{(num)}_j + x^{(num)}_j \cdot W^{(num)}_j && \in \R^d, \\
    &T^{(cat)}_j  = b^{(cat)}_j  + e_j^T W^{(cat)}_j             && \in \R^d, \\
    &T            = \mathtt{stack} \left[ T^{(num)}_1,\ \ldots,\ T^{(num)}_{k^{(num)}},\ T^{(cat)}_1,\ \ldots,\ T^{(cat)}_{k^{(cat)}} \right]              && \in \R^{k \times d}.
\end{alignat*}
where $e_j^T$ is a one-hot vector for the corresponding categorical feature.

\textbf{Transformer.} At this stage, the embedding of the \CLS\ token (or ``classification token'', or ``output token'', see \citet{bert}) is appended to $T$ and $L$ Transformer layers $F_1,\ \dotsc,\ F_L$ are applied:
$$
    T_0 = \mathtt{stack} \lrb{\CLS,\ T} \qquad T_i = F_i(T_{i - 1}).
$$
We use the PreNorm variant for easier optimization \citep{prenorm}, see \autoref{fig:blocks}. In the PreNorm setting, we also found it to be necessary to remove the first normalization from the first Transformer layer to achieve good performance. See the original paper \citep{transformer} for the background on Multi-Head Self-Attention (MHSA) and the Feed Forward module. See supplementary for details such as activations, placement of normalizations and dropout modules \citep{dropout}.

\textbf{Prediction.} The final representation of the \CLS\ token is used for prediction:
$$
    \hat{y} = \Linear(\ReLU(\LayerNorm(T^{\CLS}_{L}))).
$$

\textbf{Limitations.} \architecture\ requires more resources (both hardware and time) for training than simple models such as ResNet and may not be easily scaled to datasets when the number of features is ``too large'' (it is determined by the available hardware and time budget).
Consequently, widespread usage of \architecture\ for solving tabular data problems can lead to greater CO2 emissions produced by ML pipelines, since tabular data problems are ubiquitous.
The main cause of the described problem lies in the quadratic complexity of the vanilla MHSA with respect to the number of features.
However, the issue can be alleviated by using efficient approximations of MHSA \citep{efficient-transformers}.
Additionally, it is still possible to distill \architecture\ into simpler architectures for better inference performance.
We report training times and the used hardware in supplementary.

\subsection{Other models}
\label{sec:other-models}

In this section, we list the existing models designed specifically for tabular data that we include in the comparison.

\begin{itemize}[nosep]
    \item \textbf{SNN} \citep{snn}. An MLP-like architecture with the SELU activation that enables training deeper models.
    \item \textbf{NODE} \citep{node}. A differentiable ensemble of oblivious decision trees.
    \item \textbf{TabNet} \citep{tabnet}. A recurrent architecture that alternates dynamical reweighing of features and conventional feed-forward modules.
    \item \textbf{GrowNet} \citep{grownet}. Gradient boosted weak MLPs. The official implementation supports only classification and regression problems.
    \item \textbf{DCN V2} \citep{dcn2}. Consists of an MLP-like module and the feature crossing module (a combination of linear layers and multiplications).
    \item \textbf{AutoInt} \citep{autoint}. Transforms features to embeddings and applies a series of attention-based transformations to the embeddings.
    \item \textbf{XGBoost} \citep{xgboost}. One of the most popular GBDT implementations.
    \item \textbf{CatBoost} \citep{catboost}. GBDT implementation that uses oblivious decision trees \citep{odt} as weak learners.
\end{itemize}

\section{Experiments}
In this section, we compare DL models to each other as well as to GBDT.
Note that in the main text, we report only the key results.
In supplementary, we provide: (1) the results for all models on all datasets; (2) information on hardware; (3) training times for ResNet and \architecture.

\subsection{Scope of the comparison}
In our work, we focus on the relative performance of different architectures and do not employ various model-agnostic DL practices, such as pretraining, additional loss functions, data augmentation, distillation, learning rate warmup, learning rate decay and many others.
While these practices can potentially improve the performance, our goal is to evaluate the impact of inductive biases imposed by the different model architectures.

\subsection{Datasets}
\label{sec:datasets}

We use a diverse set of eleven public datasets (see supplementary for the detailed description). For each dataset, there is exactly one train-validation-test split, so all algorithms use the same splits. The datasets include: California Housing (CA, real estate data, \citet{california}), Adult (AD, income estimation, \citet{adult}), \mbox{Helena (HE, anonymized dataset, \citet{helena-jannis})}, \mbox{Jannis (JA, anonymized dataset, \citet{helena-jannis})}, Higgs (HI, simulated physical particles, \citet{higgs}; we use the version with 98K samples available at the OpenML repository \citep{openml}), ALOI (AL, images, \citet{aloi}), Epsilon (EP, simulated physics experiments), Year (YE, audio features, \citet{year}), Covertype (CO, forest characteristics, \citet{covertype}), Yahoo (YA, search queries, \citet{yahoo}), Microsoft (MI, search queries, \citet{microsoft}). We follow the pointwise approach to learning-to-rank and treat ranking problems (Microsoft, Yahoo) as regression problems. The dataset properties are summarized in \autoref{tab:datasets}.

\begin{table}[ht]
    \setlength\tabcolsep{2.2pt}
    \centering
    \caption{Dataset properties. Notation: \mbox{``RMSE'' \textasciitilde\ root-mean-square error}, \mbox{``Acc.'' \textasciitilde\ accuracy}.}
    \label{tab:datasets}
    \vspace{1em}
    {\footnotesize \begin{tabular}{lccccccccccc}
\toprule
{} & CA & AD & HE & JA & HI & AL & EP & YE & CO & YA & MI \\
\midrule
\#objects & 20640 & 48842 & 65196 & 83733 & 98050 & 108000 & 500000 & 515345 & 581012 & 709877 & 1200192 \\
\#num. features & 8 & 6 & 27 & 54 & 28 & 128 & 2000 & 90 & 54 & 699 & 136 \\
\#cat. features & 0 & 8 & 0 & 0 & 0 & 0 & 0 & 0 & 0 & 0 & 0 \\
metric & RMSE & Acc. & Acc. & Acc. & Acc. & Acc. & Acc. & RMSE & Acc. & RMSE & RMSE \\
\#classes & -- & 2 & 100 & 4 & 2 & 1000 & 2 & -- & 7 & -- & -- \\
\bottomrule
\end{tabular}}
\end{table}

\subsection{Implementation details}
\label{sec:implementation-details}

\textbf{Data preprocessing}.
Data preprocessing is known to be vital for DL models.
For each dataset, the same preprocessing was used for all deep models for a fair comparison.
By default, we used the quantile transformation from the Scikit-learn library \citep{scikit-learn}.
We apply standardization (mean subtraction and scaling) to Helena and ALOI.
The latter one represents image data, and standardization is a common practice in computer vision.
On the Epsilon dataset, we observed preprocessing to be detrimental to deep models’ performance, so we use the raw features on this dataset.
We apply standardization to regression targets for all algorithms.

\textbf{Tuning}.
For every dataset, we carefully tune each model's hyperparameters.
The best hyperparameters are the ones that perform best on the validation set, so the test set is never used for tuning.
For most algorithms, we use the Optuna library \citep{optuna} to run Bayesian optimization (the Tree-Structured Parzen Estimator algorithm), which is reported to be superior to random search \citep{hp-tuning}.
For the rest, we iterate over predefined sets of configurations recommended by corresponding papers.
We provide parameter spaces and grids in supplementary.
We set the budget for Optuna-based tuning in terms of \textit{iterations} and provide additional analysis on setting the budget in terms of \textit{time} in supplementary.

\textbf{Evaluation}. For each tuned configuration, we run 15 experiments with different random seeds and report the performance on the test set. For some algorithms, we also report the performance of default configurations without hyperparameter tuning.

\textbf{Ensembles}. For each model, on each dataset, we obtain three ensembles by splitting the 15 single models into three disjoint groups of equal size and averaging predictions of single models within each group.

\textbf{Neural networks}. We minimize cross-entropy for classification problems and mean squared error for regression problems. For TabNet and GrowNet, we follow the original implementations and use the Adam optimizer \citep{adam}. For all other algorithms, we use the AdamW optimizer \citep{adamw}. We do not apply learning rate schedules. For each dataset, we use a predefined batch size for all algorithms unless special instructions on batch sizes are given in the corresponding papers (see supplementary). We continue training until there are $\texttt{patience} + 1$ consecutive epochs without improvements on the validation set; we set $\texttt{patience} = 16$ for all algorithms.

\textbf{Categorical features.} For XGBoost, we use one-hot encoding. For CatBoost, we employ the built-in support for categorical features. For Neural Networks, we use embeddings of the same dimensionality for all categorical features.

\subsection{Comparing DL models}

\begin{table}[ht]
    \setlength\tabcolsep{2.2pt}
    \centering
    \caption{Results for DL models. The metric values averaged over 15 random seeds are reported. See supplementary for standard deviations.
    For each dataset, top results are in \textbf{bold}.
    ``Top'' means ``the gap between this result and the result with the best score is not statistically significant''.
    For each dataset, ranks are calculated by sorting the reported scores; the ``rank'' column reports the average rank across all datasets.
    Notation:
    \mbox{FT-T \textasciitilde\ \architecture},
    \mbox{\textdownarrow\ \textasciitilde\ RMSE},
    \mbox{\textuparrow\ \textasciitilde\ accuracy}}
    \label{tab:neural-networks}
    \vspace{1em}
    {\footnotesize \begin{tabular}{lccccccccccc|c}
\toprule
{} & CA \textdownarrow & AD \textuparrow & HE \textuparrow & JA \textuparrow & HI \textuparrow & AL \textuparrow & EP \textuparrow & YE \textdownarrow & CO \textuparrow & YA \textdownarrow & MI \textdownarrow & rank (std) 
\\
\midrule
TabNet          & $0.510$ & $0.850$ & $0.378$ & $0.723$ & $0.719$ & $0.954$ & $0.8896$ & $8.909$ & $0.957$ & $0.823$ & $0.751$ & $7.5$ $(2.0)$
\\
SNN             & $0.493$ & $0.854$ & $0.373$ & $0.719$ & $0.722$ & $0.954$ & $0.8975$ & $8.895$ & $0.961$ & $0.761$ & $0.751$ & $6.4$ $(1.4)$
\\
AutoInt         & $0.474$ & $\mathbf{0.859}$ & $0.372$ & $0.721$ & $0.725$ & $0.945$ & $0.8949$ & $8.882$ & $0.934$ & $0.768$ & $0.750$ & $5.7$ $(2.3)$
\\
GrowNet         & $0.487$ & $\mathbf{0.857}$ & -- & -- & $0.722$ & -- & $0.8970$ & $8.827$ & -- & $0.765$ & $0.751$ & $5.7$ $(2.2)$
\\
MLP             & $0.499$ & $0.852$ & $0.383$ & $0.719$ & $0.723$ & $0.954$ & $0.8977$ & $8.853$ & $0.962$ & $0.757$ & $0.747$ & $4.8$ $(1.9)$
\\
DCN2            & $0.484$ & $0.853$ & $0.385$ & $0.716$ & $0.723$ & $0.955$ & $0.8977$ & $8.890$ & $0.965$ & $0.757$ & $0.749$ & $4.7$ $(2.0)$
\\
NODE            & $0.464$ & $\mathbf{0.858}$ & $0.359$ & $0.727$ & $0.726$ & $0.918$ & $0.8958$ & $\mathbf{8.784}$ & $0.958$ & $\mathbf{0.753}$ & $\mathbf{0.745}$ & $3.9$ $(2.8)$
\\
ResNet          & $0.486$ & $0.854$ & $\mathbf{0.396}$ & $0.728$ & $0.727$ & $\mathbf{0.963}$ & $0.8969$ & $8.846$ & $0.964$ & $0.757$ & $0.748$ & $3.3$ $(1.8)$
\\
FT-T            & $\mathbf{0.459}$ & $\mathbf{0.859}$ & $0.391$ & $\mathbf{0.732}$ & $\mathbf{0.729}$ & $0.960$ & $\mathbf{0.8982}$ & $8.855$ & $\mathbf{0.970}$ & $0.756$ & $0.746$ & $1.8$ $(1.2)$
\\
\bottomrule
\end{tabular}}
\end{table}

\autoref{tab:neural-networks} reports the results for deep architectures.
\\\textbf{The main takeaways}:
\begin{itemize}[nosep]
    \item MLP is still a good sanity check
    \item ResNet turns out to be an effective baseline that none of the competitors can consistently outperform.
    \item FT-Transformer performs best on most tasks and becomes a new powerful solution for the field.
    \item Tuning makes simple models such as MLP and ResNet competitive, so we recommend tuning baselines when possible. Luckily, today, it is more approachable with libraries such as Optuna \citep{optuna}.
\end{itemize}

Among other models, NODE \citep{node} is the only one that demonstrates high performance on several tasks.
However, it is still inferior to ResNet on six datasets (Helena, Jannis, Higgs, ALOI, Epsilon, Covertype), while being a more complex solution.
Moreover, it is not a truly ``single'' model; in fact, it often contains significantly more parameters than ResNet and \architecture\ and has an ensemble-like structure.
We illustrate that by comparing \textit{ensembles} in \autoref{tab:node}.
The results indicate that \architecture\ and ResNet benefit more from ensembling; in this regime, \architecture\ outperforms NODE and the gap between ResNet and NODE is significantly reduced.
Nevertheless, NODE remains a prominent solution among tree-based approaches.

\begin{table}[htbp]
    \setlength\tabcolsep{2.2pt}
    \centering
    \caption{Results for ensembles of DL models with the highest ranks (see \autoref{tab:neural-networks}). For each \mbox{model-dataset} pair, the metric value averaged over three ensembles is reported.
    See supplementary for standard deviations.
    Depending on the dataset, the highest accuracy or the lowest RMSE is in \textbf{bold}.
    Due to the limited precision, some \textit{different} values are represented with the same figures.
    Notation:
    \mbox{\textdownarrow\ \textasciitilde\ RMSE},
    \mbox{\textuparrow\ \textasciitilde\ accuracy}.}
    \label{tab:node}
    \vspace{1em}
    {\footnotesize \begin{tabular}{lccccccccccc}
\toprule
{} & CA \textdownarrow & AD \textuparrow & HE \textuparrow & JA \textuparrow & HI \textuparrow & AL \textuparrow & EP \textuparrow & YE \textdownarrow & CO \textuparrow & YA \textdownarrow & MI \textdownarrow \\
\midrule
NODE            & $0.461$ & $0.860$ & $0.361$ & $0.730$ & $0.727$ & $0.921$ & $0.8970$ & $\mathbf{8.716}$ & $0.965$ & $0.750$ & $0.744$ \\
ResNet          & $0.478$ & $0.857$ & $0.398$ & $0.734$ & $0.731$ & $0.966$ & $0.8976$ & $8.770$ & $0.967$ & $0.751$ & $0.745$ \\
FT-Transformer  & $\mathbf{0.448}$ & $\mathbf{0.860}$ & $\mathbf{0.398}$ & $\mathbf{0.739}$ & $\mathbf{0.731}$ & $\mathbf{0.967}$ & $\mathbf{0.8984}$ & $8.751$ & $\mathbf{0.973}$ & $\mathbf{0.747}$ & $\mathbf{0.743}$ \\
\bottomrule
\end{tabular}
}
\end{table}

\subsection{Comparing DL models and GBDT}
\label{sec:nn-gbdt}

In this section, our goal is to check whether DL models are \textit{conceptually} ready to outperform GBDT.
To this end, we compare the best possible metric values that one can achieve using GBDT or DL models, without taking speed and hardware requirements into account (undoubtedly, GBDT is a more lightweight solution).
We accomplish that by comparing \textit{ensembles} instead of single models since GBDT is essentially an ensembling technique and we expect that deep architectures will benefit more from ensembling \citep{ensembles-loss}.
We report the results in \autoref{tab:nn-gbdt}.

\begin{table}[htbp]
    \setlength\tabcolsep{2.2pt}
    \centering
    \caption{Results for ensembles of GBDT and the main DL models. For each model-dataset pair, the metric value averaged over three ensembles is reported.
    See supplementary for standard deviations.
    Notation follows \autoref{tab:node}.}
    \label{tab:nn-gbdt}
    \vspace{1em}
    {\footnotesize \begin{tabular}{lccccccccccc}
\toprule
{} & CA \textdownarrow & AD \textuparrow & HE \textuparrow & JA \textuparrow & HI \textuparrow & AL \textuparrow & EP \textuparrow & YE \textdownarrow & CO \textuparrow & YA \textdownarrow & MI \textdownarrow \\
\midrule
\multicolumn{12}{c}{Default hyperparameters}\\
\midrule
XGBoost         & $0.462$ & $\mathbf{0.874}$ & $0.348$ & $0.711$ & $0.717$ & $0.924$ & $0.8799$ & $9.192$ & $0.964$ & $0.761$ & $0.751$ \\
CatBoost        & $\mathbf{0.428}$ & $0.873$ & $0.386$ & $0.724$ & $0.728$ & $0.948$ & $0.8893$ & $8.885$ & $0.910$ & $0.749$ & $0.744$ \\
FT-Transformer  & $0.454$ & $0.860$ & $\mathbf{0.395}$ & $\mathbf{0.734}$ & $\mathbf{0.731}$ & $\mathbf{0.966}$ & $\mathbf{0.8969}$ & $\mathbf{8.727}$ & $\mathbf{0.973}$ & $\mathbf{0.747}$ & $\mathbf{0.742}$ \\
\midrule
\multicolumn{12}{c}{Tuned hyperparameters}\\
\midrule
XGBoost         & $0.431$ & $0.872$ & $0.377$ & $0.724$ & $0.728$ & -- & $0.8861$ & $8.819$ & $0.969$ & $\mathbf{0.732}$ & $0.742$ \\
CatBoost        & $\mathbf{0.423}$ & $\mathbf{0.874}$ & $0.388$ & $0.727$ & $0.729$ & -- & $0.8898$ & $8.837$ & $0.968$ & $0.740$ & $\mathbf{0.741}$ \\
ResNet          & $0.478$ & $0.857$ & $0.398$ & $0.734$ & $0.731$ & $0.966$ & $0.8976$ & $8.770$ & $0.967$ & $0.751$ & $0.745$ \\
FT-Transformer  & $0.448$ & $0.860$ & $\mathbf{0.398}$ & $\mathbf{0.739}$ & $\mathbf{0.731}$ & $\mathbf{0.967}$ & $\mathbf{0.8984}$ & $\mathbf{8.751}$ & $\mathbf{0.973}$ & $0.747$ & $0.743$ \\
\bottomrule
\end{tabular}}
\end{table}

\textbf{Default hyperparameters}.
We start with the default configurations to check the ``out-of-the-box'' performance, which is an important practical scenario.
The default \architecture\ implies a configuration with all hyperparameters set to some specific values that we provide in supplementary.
\autoref{tab:nn-gbdt} demonstrates that the ensemble of \architecture s mostly outperforms the ensembles of GBDT, which is not the case for only two datasets (California Housing, Adult).
Interestingly, the ensemble of default \architecture s performs quite on par with the ensembles of tuned \architecture s.
\\\textbf{The main takeaway}: \architecture\ allows building powerful ensembles out of the box.

\textbf{Tuned hyperparameters}.
Once hyperparameters are properly tuned, GBDTs start dominating on some datasets (California Housing, Adult, Yahoo; see \autoref{tab:nn-gbdt}).
In those cases, the gaps are significant enough to conclude that DL models do not universally outperform GBDT.
Importantly, the fact that DL models outperform GBDT on most of the tasks does \textit{not} mean that DL solutions are ``better'' in any sense.
In fact, it only means that the constructed benchmark is slightly biased towards ``DL-friendly'' problems.
Admittedly, GBDT remains an unsuitable solution to multiclass problems with a large number of classes.
Depending on the number of classes, GBDT can demonstrate unsatisfactory performance (Helena) or even be untunable due to extremely slow training (ALOI).
\\\textbf{The main takeaways}:
\begin{itemize}[nosep]
    \item there is still no universal solution among DL models and GBDT
    \item DL research efforts aimed at surpassing GBDT should focus on datasets where GBDT outperforms state-of-the-art DL solutions. Note that including “DL-friendly” problems is still important to avoid degradation on such problems.
\end{itemize}

\subsection{An intriguing property of \architecture}
\label{sec:intriguing-property}
\autoref{tab:nn-gbdt} tells one more important story.
Namely, \architecture\ delivers most of its advantage over the ``conventional'' DL model in the form of ResNet exactly on those problems where GBDT is superior to ResNet (California Housing, Adult, Covertype, Yahoo, Microsoft) while performing on par with ResNet on the remaining problems.
In other words, \architecture\ provides competitive performance on all tasks, while GBDT and ResNet perform well only on some subsets of the tasks.
This observation may be the evidence that \architecture\ is a more ``universal'' model for tabular data problems.
We develop this intuition further in \autoref{sec:synthetic}.
Note that the described phenomenon is not related to ensembling and is observed for single models too (see supplementary).

\section{Analysis}

\subsection{When \architecture\ is better than ResNet?}
\label{sec:synthetic}

In this section, we make the first step towards understanding the difference in behavior between \architecture\ and ResNet, which was first observed in \autoref{sec:intriguing-property}.
To achieve that, we design a sequence of synthetic tasks where the difference in performance of the two models gradually changes from negligible to dramatic.
Namely, we generate and \textit{fix} objects $\{x_i\}_{i=1}^n$, perform the train-val-test split \textit{once} and interpolate between two regression targets: $f_{GBDT}$, which is supposed to be easier for GBDT and $f_{DL}$, which is expected to be easier for ResNet. Formally, for one object:

$$
    x \sim \mathcal{N}(0, I_k), \qquad y = \alpha \cdot f_{GBDT}(x) + (1 - \alpha) \cdot f_{DL}(x).
$$

\begin{wrapfigure}{htbp}{0.5\textwidth}
    \vspace{-30pt}
    \begin{center}
        \scalebox{0.55}{\input{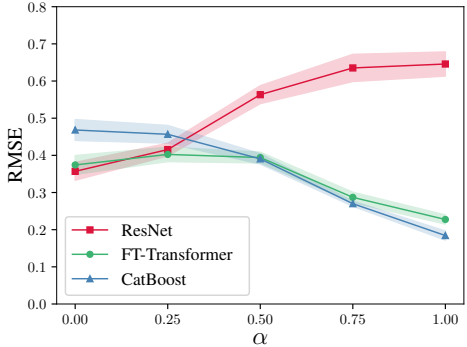}}
    \end{center}
    \caption{Test RMSE averaged over five seeds (shadows represent std. dev.). One $\alpha$ corresponds to one task; each task has the same set of train, validation and test features, but different targets.}
    \label{fig:synthetic}
\end{wrapfigure}

where $f_{GBDT}(x)$ is an average prediction of 30 randomly constructed decision trees, and $f_{DL}(x)$ is an MLP with three randomly initialized hidden layers. Both $f_{GBDT}$ and $f_{DL}$ are generated once, i.e. the same functions are applied to all objects (see supplementary for details). The resulting targets are standardized before training. The results are visualized in \autoref{fig:synthetic}. ResNet and \architecture\ perform similarly well on the ResNet-friendly tasks and outperform CatBoost on those tasks. However, the ResNet's relative performance drops significantly when the target becomes more GBDT friendly. By contrast, \architecture\ yields competitive performance across the whole range of tasks.

The conducted experiment reveals a type of functions that are better approximated by \architecture\ than by ResNet. Additionally, the fact that these functions are based on decision trees correlates with the observations in \autoref{sec:intriguing-property} and the results in \autoref{tab:nn-gbdt}, where \architecture\ shows the most convincing improvements over ResNet exactly on those datasets where GBDT outperforms ResNet.

\subsection{Ablation study}
\label{sec:ablation}

In this section, we test some design choices of \architecture.

First, we compare \architecture\ with AutoInt \citep{autoint}, since it is the closest competitor in its spirit. AutoInt also converts all features to embeddings and applies self-attention on top of them. However, in its details, AutoInt significantly differs from \architecture: its embedding layer does not include feature biases, its backbone significantly differs from the vanilla Transformer \citep{transformer}, and the inference mechanism does not use the \CLS\ token.

Second, we check whether feature biases in \tokenizer\ are essential for good performance.

We tune and evaluate \architecture\ without feature biases following the same protocol as in \autoref{sec:implementation-details} and reuse the remaining numbers from \autoref{tab:neural-networks}. The results averaged over 15 runs are reported in \autoref{tab:ablation} and demonstrate both the superiority of the Transformer's backbone to that of AutoInt and the necessity of feature biases.

\begin{table}[h]
\caption{The results of the comparison between \architecture\ and two attention-based alternatives: AutoInt and \architecture\ without feature biases. Notation follows \autoref{tab:neural-networks}.}
\vspace{1em}
\label{tab:ablation}
\centering
\setlength\tabcolsep{2.4pt}
{\small \begin{tabular}{lcccccccc}
\toprule
{} & CA \textdownarrow & HE \textuparrow & JA \textuparrow & HI \textuparrow & AL \textuparrow & YE \textdownarrow & CO \textuparrow & MI \textdownarrow \\
\midrule
AutoInt         & $0.474$ & $0.372$ & $0.721$ & $0.725$ & $0.945$ & $8.882$ & $0.934$ & $0.750$ \\
FT-Transformer (w/o feature biases) & $0.470$ & $0.381$ & $0.724$ & $\mathbf{0.727}$ & $0.958$ & $\mathbf{8.843}$ & $0.964$ & $0.751$ \\
FT-Transformer  & $\mathbf{0.459}$ & $\mathbf{0.391}$ & $\mathbf{0.732}$ & $\mathbf{0.729}$ & $\mathbf{0.960}$ & $\mathbf{8.855}$ & $\mathbf{0.970}$ & $\mathbf{0.746}$ \\
\bottomrule
\end{tabular}}
\end{table}

\subsection{Obtaining feature importances from attention maps}

In this section, we evaluate attention maps as a source of information on feature importances for \architecture\ for a given set of samples. For the $i$-th sample, we calculate the average attention map $p_i$ for the \CLS\ token from Transformer's forward pass. Then, the obtained individual distributions are averaged into one distribution $p$ that represents the feature importances:
$$
    p = \frac{1}{n_{samples}} \sum_i p_i \qquad p_i = \frac{1}{n_{heads} \times L} \sum_{h,l} p_{ihl}.
$$
where $p_{ihl}$ is the $h$-th head's attention map for the \CLS\ token from the forward pass of the $l$-th layer on the $i$-th sample. The main advantage of the described heuristic technique is its efficiency: it requires a single forward for one sample.

In order to evaluate our approach, we compare it with Integrated Gradients (IG, \citet{integrated-gradients}), a general technique applicable to any differentiable model. We use permutation test (PT, \citet{random-forest}) as a reasonable interpretable method that allows us to establish a constructive metric, namely, rank correlation. We run all the methods on the train set and summarize results in \autoref{tab:feature-importances}. Interestingly, the proposed method yields reasonable feature importances and performs similarly to IG (note that this does not imply similarity to IG's feature importances). Given that IG can be orders of magnitude slower and the ``baseline'' in the form of PT requires $(n_{features} + 1)$ forward passes (versus one for the proposed method), we conclude that the simple averaging of attention maps can be a good choice in terms of cost-effectiveness.

\begin{table}[h]
\setlength\tabcolsep{2.4pt}
\centering
\caption{Rank correlation (takes values in $[-1,\ 1]$) between permutation test's feature importances ranking and two alternative rankings: Attention Maps (AM) and Integrated Gradients (IG). Means and standard deviations over five runs are reported.}
\label{tab:feature-importances}
\vspace{1em}
{\small \begin{tabular}{ccccccccc}
\toprule
   & CA & HE & JA & HI & AL & YE & CO & MI \\
\midrule
AM & $0.81\ (0.05)$ & $0.77\ (0.03)$ & $0.78\ (0.05)$ & $0.91\ (0.03)$ & $0.84\ (0.01)$ & $0.92\ (0.01)$ & $0.84\ (0.04)$ & $0.86\ (0.02)$ \\
IG & $0.84\ (0.08)$ & $0.74\ (0.03)$ & $0.75\ (0.04)$ & $0.72\ (0.03)$ & $0.89\ (0.01)$ & $0.50\ (0.03)$ & $0.90\ (0.02)$ & $0.56\ (0.02)$ \\
\bottomrule
\end{tabular}
}
\end{table}

\section{Conclusion}

In this work, we have investigated the status quo in the field of deep learning for tabular data and improved the state of baselines in tabular DL.
First, we have demonstrated that a simple ResNet-like architecture can serve as an effective baseline.
Second, we have proposed FT-Transformer --- a simple adaptation of the Transformer architecture that outperforms other DL solutions on most of the tasks.
We have also compared the new baselines with GBDT and demonstrated that GBDT still dominates on some tasks.
The code and all the details of the study are open-sourced \footnote{\url{\repository}}, and we hope that our evaluation and two simple models (ResNet and \architecture) will serve as a basis for further developments on tabular DL.

\medskip

\bibliographystyle{abbrvnat}
\bibliography{references}

\newpage
\renewcommand\thesubsection{\Alph{subsection}}

\section*{Supplementary material}

\subsection{Software and hardware}
For most model-dataset pairs the workflow was as follows:
\begin{itemize}[nosep]
    \item tune the model on any suitable hardware
    \item evaluate the tuned model on one or more NVidia Tesla V100 32Gb
\end{itemize}
All the experiments were conducted under the same conditions in terms of software versions.
For almost all experiments the used hardware can be found in the source code.

\subsection{Data}

\subsubsection{Datasets}
\label{sec:S-datasets}

\begin{table}[h]
    \setlength\tabcolsep{2.4pt}
    \centering
    \caption{Datasets description}
    \label{tab:S-datasets}
    \vspace{1em}
    \begin{tabular}{lcccccccc}
    \toprule
    Name & Abbr & \# Train & \# Validation & \# Test & \# Num & \# Cat & Task type & Batch size \\
    \midrule

    California Housing & CA & $13209$ & $3303$ & $4128$ & $8$ & $0$ & Regression & 256 \\
    Adult & AD & $26048$ & $6513$ & $16281$ & $6$ & $8$ & Binclass & 256 \\
    Helena & HE & $41724$ & $10432$ & $13040$ & $27$ & $0$ & Multiclass & 512 \\
    Jannis & JA & $53588$ & $13398$ & $16747$ & $54$ & $0$ & Multiclass & 512 \\
    Higgs Small & HI & $62752$ & $15688$ & $19610$ & $28$ & $0$ & Binclass & 512 \\
    ALOI & AL & $69120$ & $17280$ & $21600$ & $128$ & $0$ & Multiclass & 512 \\
    Epsilon & EP & $320000$ & $80000$ & $100000$ & $2000$ & $0$ & Binclass & 1024 \\
    Year & YE & $370972$ & $92743$ & $51630$ & $90$ & $0$ & Regression & 1024 \\
    Covtype & CO & $371847$ & $92962$ & $116203$ & $54$ & $0$ & Multiclass & 1024 \\
    Yahoo & YA & $473134$ & $71083$ & $165660$ & $699$ & $0$ & Regression & 1024 \\
    Microsoft & MI & $723412$ & $235259$ & $241521$ & $136$ & $0$ & Regression & 1024 \\

    \bottomrule
\end{tabular}

\end{table}

\subsubsection{Preprocessing}
For regression problems, we standardize the target values:
\begin{equation}
    y_{new} = \frac{y_{old} - \mathtt{mean}(y_{train}))}{\mathtt{std}(y_{train})}
\end{equation}

The feature preprocessing for DL models is described in the main text. Note that we add noise from $\mathcal{N}(0, 1e-3)$ to train numerical features for calculating the parameters (quantiles) of the quantile preprocessing as a workaround for features with few distinct values (see the source code for the exact implementation). The preprocessing is then applied to \textit{original} features. We do not preprocess features for GBDTs, since this family of algorithms is insensitive to feature shifts and scaling.

\subsection{Results for all algorithms on all datasets}

To measure statistical significance in the main text and in the tables in this section, we use the one-sided \citet{wilcoxon} test with $p = 0.01$.

\autoref{tab:S-single-models} and \autoref{tab:S-ensembles} report all results for all models on all datasets.

\begin{sidewaystable}
    \centering
    \setlength\tabcolsep{2.4pt}
    \caption{Results for single models with standard deviations. For each dataset, top results for baseline neural networks are in \textbf{bold}, top results for baseline neural networks and \architecture\ are in \textbf{\textcolor{blue}{blue}}, the overall top results are in \textbf{\textcolor{red}{red}}. “Top” means “the gap between this result and the result with the best mean score is not statistically significant”. ``d'' stands for ``default''. The remaining notation follows those from the main text. Best viewed in colors.}
    \label{tab:S-single-models}
    \small
    \begin{tabular}{lccccccccccc}
\toprule
{} & CA \textdownarrow & AD \textuparrow & HE \textuparrow & JA \textuparrow & HI \textuparrow & AL \textuparrow & EP \textuparrow & YE \textdownarrow & CO \textuparrow & YA \textdownarrow & MI \textdownarrow \\
\midrule
\multicolumn{12}{c}{Baseline Neural Networks}\\
\midrule
TabNet          & $0.510 \scriptscriptstyle \pm \scriptstyle 7.6e\text{-}3$ & $0.850 \scriptscriptstyle \pm \scriptstyle 5.2e\text{-}3$ & $0.378 \scriptscriptstyle \pm \scriptstyle 1.7e\text{-}3$ & $0.723 \scriptscriptstyle \pm \scriptstyle 3.5e\text{-}3$ & $0.719 \scriptscriptstyle \pm \scriptstyle 1.7e\text{-}3$ & $0.954 \scriptscriptstyle \pm \scriptstyle 1.0e\text{-}3$ & $0.8896 \scriptscriptstyle \pm \scriptstyle 3.1e\text{-}3$ & $8.909 \scriptscriptstyle \pm \scriptstyle 2.3e\text{-}2$ & $0.957 \scriptscriptstyle \pm \scriptstyle 7.5e\text{-}3$ & $0.823 \scriptscriptstyle \pm \scriptstyle 9.2e\text{-}3$ & $0.751 \scriptscriptstyle \pm \scriptstyle 9.4e\text{-}4$ \\
SNN             & $0.493 \scriptscriptstyle \pm \scriptstyle 4.6e\text{-}3$ & $0.854 \scriptscriptstyle \pm \scriptstyle 1.8e\text{-}3$ & $0.373 \scriptscriptstyle \pm \scriptstyle 2.8e\text{-}3$ & $0.719 \scriptscriptstyle \pm \scriptstyle 1.6e\text{-}3$ & $0.722 \scriptscriptstyle \pm \scriptstyle 2.2e\text{-}3$ & $0.954 \scriptscriptstyle \pm \scriptstyle 1.6e\text{-}3$ & $\mathbf{0.8975 \scriptscriptstyle \pm \scriptstyle 2.4e\text{-}4}$ & $8.895 \scriptscriptstyle \pm \scriptstyle 1.9e\text{-}2$ & $0.961 \scriptscriptstyle \pm \scriptstyle 2.0e\text{-}3$ & $0.761 \scriptscriptstyle \pm \scriptstyle 5.3e\text{-}4$ & $0.751 \scriptscriptstyle \pm \scriptstyle 5.2e\text{-}4$ \\
AutoInt         & $0.474 \scriptscriptstyle \pm \scriptstyle 3.3e\text{-}3$ & $\mathbf{\textcolor{blue}{0.859 \scriptscriptstyle \pm \scriptstyle 1.5e\text{-}3}}$ & $0.372 \scriptscriptstyle \pm \scriptstyle 2.5e\text{-}3$ & $0.721 \scriptscriptstyle \pm \scriptstyle 2.3e\text{-}3$ & $\mathbf{0.725 \scriptscriptstyle \pm \scriptstyle 1.7e\text{-}3}$ & $0.945 \scriptscriptstyle \pm \scriptstyle 1.3e\text{-}3$ & $0.8949 \scriptscriptstyle \pm \scriptstyle 5.8e\text{-}4$ & $8.882 \scriptscriptstyle \pm \scriptstyle 3.3e\text{-}2$ & $0.934 \scriptscriptstyle \pm \scriptstyle 3.5e\text{-}3$ & $0.768 \scriptscriptstyle \pm \scriptstyle 1.1e\text{-}3$ & $0.750 \scriptscriptstyle \pm \scriptstyle 6.1e\text{-}4$ \\
GrowNet         & $0.487 \scriptscriptstyle \pm \scriptstyle 7.1e\text{-}3$ & $\mathbf{\textcolor{blue}{0.857 \scriptscriptstyle \pm \scriptstyle 1.9e\text{-}3}}$ & -- & -- & $0.722 \scriptscriptstyle \pm \scriptstyle 1.6e\text{-}3$ & -- & $0.8970 \scriptscriptstyle \pm \scriptstyle 5.7e\text{-}4$ & $8.827 \scriptscriptstyle \pm \scriptstyle 3.8e\text{-}2$ & -- & $0.765 \scriptscriptstyle \pm \scriptstyle 1.2e\text{-}3$ & $0.751 \scriptscriptstyle \pm \scriptstyle 4.7e\text{-}4$ \\
MLP             & $0.499 \scriptscriptstyle \pm \scriptstyle 2.9e\text{-}3$ & $0.852 \scriptscriptstyle \pm \scriptstyle 1.9e\text{-}3$ & $0.383 \scriptscriptstyle \pm \scriptstyle 2.6e\text{-}3$ & $0.719 \scriptscriptstyle \pm \scriptstyle 1.3e\text{-}3$ & $0.723 \scriptscriptstyle \pm \scriptstyle 1.8e\text{-}3$ & $0.954 \scriptscriptstyle \pm \scriptstyle 1.4e\text{-}3$ & $\mathbf{0.8977 \scriptscriptstyle \pm \scriptstyle 4.1e\text{-}4}$ & $8.853 \scriptscriptstyle \pm \scriptstyle 3.1e\text{-}2$ & $0.962 \scriptscriptstyle \pm \scriptstyle 1.1e\text{-}3$ & $0.757 \scriptscriptstyle \pm \scriptstyle 3.5e\text{-}4$ & $0.747 \scriptscriptstyle \pm \scriptstyle 3.3e\text{-}4$ \\
DCN2            & $0.484 \scriptscriptstyle \pm \scriptstyle 2.4e\text{-}3$ & $0.853 \scriptscriptstyle \pm \scriptstyle 3.9e\text{-}3$ & $0.385 \scriptscriptstyle \pm \scriptstyle 3.0e\text{-}3$ & $0.716 \scriptscriptstyle \pm \scriptstyle 1.5e\text{-}3$ & $0.723 \scriptscriptstyle \pm \scriptstyle 1.3e\text{-}3$ & $0.955 \scriptscriptstyle \pm \scriptstyle 1.2e\text{-}3$ & $\mathbf{0.8977 \scriptscriptstyle \pm \scriptstyle 2.6e\text{-}4}$ & $8.890 \scriptscriptstyle \pm \scriptstyle 2.8e\text{-}2$ & $\mathbf{0.965 \scriptscriptstyle \pm \scriptstyle 1.0e\text{-}3}$ & $0.757 \scriptscriptstyle \pm \scriptstyle 1.9e\text{-}3$ & $0.749 \scriptscriptstyle \pm \scriptstyle 5.8e\text{-}4$ \\
NODE            & $\mathbf{0.464 \scriptscriptstyle \pm \scriptstyle 1.5e\text{-}3}$ & $\mathbf{\textcolor{blue}{0.858 \scriptscriptstyle \pm \scriptstyle 1.6e\text{-}3}}$ & $0.359 \scriptscriptstyle \pm \scriptstyle 2.0e\text{-}3$ & $\mathbf{0.727 \scriptscriptstyle \pm \scriptstyle 1.6e\text{-}3}$ & $\mathbf{0.726 \scriptscriptstyle \pm \scriptstyle 1.3e\text{-}3}$ & $0.918 \scriptscriptstyle \pm \scriptstyle 5.4e\text{-}3$ & $0.8958 \scriptscriptstyle \pm \scriptstyle 4.7e\text{-}4$ & $\mathbf{\textcolor{red}{8.784 \scriptscriptstyle \pm \scriptstyle 1.6e\text{-}2}}$ & $0.958 \scriptscriptstyle \pm \scriptstyle 1.1e\text{-}3$ & $\mathbf{\textcolor{blue}{0.753 \scriptscriptstyle \pm \scriptstyle 2.5e\text{-}4}}$ & $\mathbf{\textcolor{blue}{0.745 \scriptscriptstyle \pm \scriptstyle 2.0e\text{-}4}}$ \\
ResNet          & $0.486 \scriptscriptstyle \pm \scriptstyle 2.9e\text{-}3$ & $0.854 \scriptscriptstyle \pm \scriptstyle 1.7e\text{-}3$ & $\mathbf{\textcolor{red}{0.396 \scriptscriptstyle \pm \scriptstyle 1.7e\text{-}3}}$ & $\mathbf{0.728 \scriptscriptstyle \pm \scriptstyle 1.5e\text{-}3}$ & $\mathbf{0.727 \scriptscriptstyle \pm \scriptstyle 1.7e\text{-}3}$ & $\mathbf{\textcolor{red}{0.963 \scriptscriptstyle \pm \scriptstyle 7.5e\text{-}4}}$ & $0.8969 \scriptscriptstyle \pm \scriptstyle 4.4e\text{-}4$ & $8.846 \scriptscriptstyle \pm \scriptstyle 2.4e\text{-}2$ & $\mathbf{0.964 \scriptscriptstyle \pm \scriptstyle 1.1e\text{-}3}$ & $0.757 \scriptscriptstyle \pm \scriptstyle 6.2e\text{-}4$ & $0.748 \scriptscriptstyle \pm \scriptstyle 3.1e\text{-}4$ \\
\midrule
\multicolumn{12}{c}{\architecture}\\
\midrule
FT-Transformer\textsubscript{d} & $0.469 \scriptscriptstyle \pm \scriptstyle 3.8e\text{-}3$ & $0.857 \scriptscriptstyle \pm \scriptstyle 1.1e\text{-}3$ & $0.381 \scriptscriptstyle \pm \scriptstyle 2.4e\text{-}3$ & $0.725 \scriptscriptstyle \pm \scriptstyle 2.3e\text{-}3$ & $0.725 \scriptscriptstyle \pm \scriptstyle 1.8e\text{-}3$ & $0.953 \scriptscriptstyle \pm \scriptstyle 1.1e\text{-}3$ & $0.8959 \scriptscriptstyle \pm \scriptstyle 4.9e\text{-}4$ & $8.889 \scriptscriptstyle \pm \scriptstyle 4.6e\text{-}2$ & $0.967 \scriptscriptstyle \pm \scriptstyle 7.9e\text{-}4$ & $0.756 \scriptscriptstyle \pm \scriptstyle 8.2e\text{-}4$ & $0.747 \scriptscriptstyle \pm \scriptstyle 7.9e\text{-}4$ \\
FT-Transformer  & $\mathbf{\textcolor{blue}{0.459 \scriptscriptstyle \pm \scriptstyle 3.5e\text{-}3}}$ & $\mathbf{\textcolor{blue}{0.859 \scriptscriptstyle \pm \scriptstyle 1.0e\text{-}3}}$ & $0.391 \scriptscriptstyle \pm \scriptstyle 1.2e\text{-}3$ & $\mathbf{\textcolor{red}{0.732 \scriptscriptstyle \pm \scriptstyle 2.0e\text{-}3}}$ & $\mathbf{\textcolor{red}{0.729 \scriptscriptstyle \pm \scriptstyle 1.5e\text{-}3}}$ & $0.960 \scriptscriptstyle \pm \scriptstyle 1.1e\text{-}3$ & $\mathbf{\textcolor{red}{0.8982 \scriptscriptstyle \pm \scriptstyle 2.8e\text{-}4}}$ & $8.855 \scriptscriptstyle \pm \scriptstyle 3.1e\text{-}2$ & $\mathbf{\textcolor{red}{0.970 \scriptscriptstyle \pm \scriptstyle 6.6e\text{-}4}}$ & $0.756 \scriptscriptstyle \pm \scriptstyle 8.2e\text{-}4$ & $0.746 \scriptscriptstyle \pm \scriptstyle 4.9e\text{-}4$ \\
\midrule
\multicolumn{12}{c}{GBDT}\\
\midrule
CatBoost\textsubscript{d} & $\mathbf{\textcolor{red}{0.430 \scriptscriptstyle \pm \scriptstyle 7.4e\text{-}4}}$ & $0.873 \scriptscriptstyle \pm \scriptstyle 9.6e\text{-}4$ & $0.381 \scriptscriptstyle \pm \scriptstyle 1.5e\text{-}3$ & $0.721 \scriptscriptstyle \pm \scriptstyle 1.1e\text{-}3$ & $0.726 \scriptscriptstyle \pm \scriptstyle 8.0e\text{-}4$ & $0.946 \scriptscriptstyle \pm \scriptstyle 9.3e\text{-}4$ & $0.8880 \scriptscriptstyle \pm \scriptstyle 4.5e\text{-}4$ & $8.913 \scriptscriptstyle \pm \scriptstyle 5.5e\text{-}3$ & $0.908 \scriptscriptstyle \pm \scriptstyle 2.4e\text{-}4$ & $0.751 \scriptscriptstyle \pm \scriptstyle 2.0e\text{-}4$ & $0.745 \scriptscriptstyle \pm \scriptstyle 2.3e\text{-}4$ \\
CatBoost        & $\mathbf{\textcolor{red}{0.431 \scriptscriptstyle \pm \scriptstyle 1.5e\text{-}3}}$ & $0.873 \scriptscriptstyle \pm \scriptstyle 1.2e\text{-}3$ & $0.385 \scriptscriptstyle \pm \scriptstyle 1.1e\text{-}3$ & $0.723 \scriptscriptstyle \pm \scriptstyle 1.5e\text{-}3$ & $0.725 \scriptscriptstyle \pm \scriptstyle 1.5e\text{-}3$ & -- & $0.8880 \scriptscriptstyle \pm \scriptstyle 5.8e\text{-}4$ & $8.877 \scriptscriptstyle \pm \scriptstyle 6.0e\text{-}3$ & $0.966 \scriptscriptstyle \pm \scriptstyle 2.7e\text{-}4$ & $0.743 \scriptscriptstyle \pm \scriptstyle 2.4e\text{-}4$ & $0.743 \scriptscriptstyle \pm \scriptstyle 2.1e\text{-}4$ \\
XGBoost\textsubscript{d} & $0.462 \scriptscriptstyle \pm \scriptstyle 0.0$ & $\mathbf{\textcolor{red}{0.874 \scriptscriptstyle \pm \scriptstyle 0.0}}$ & $0.348 \scriptscriptstyle \pm \scriptstyle 0.0$ & $0.711 \scriptscriptstyle \pm \scriptstyle 0.0$ & $0.717 \scriptscriptstyle \pm \scriptstyle 0.0$ & $0.924 \scriptscriptstyle \pm \scriptstyle 0.0$ & $0.8799 \scriptscriptstyle \pm \scriptstyle 0.0$ & $9.192 \scriptscriptstyle \pm \scriptstyle 0.0$ & $0.964 \scriptscriptstyle \pm \scriptstyle 0.0$ & $0.761 \scriptscriptstyle \pm \scriptstyle 0.0$ & $0.751 \scriptscriptstyle \pm \scriptstyle 0.0$ \\
XGBoost         & $0.433 \scriptscriptstyle \pm \scriptstyle 1.6e\text{-}3$ & $0.872 \scriptscriptstyle \pm \scriptstyle 4.6e\text{-}4$ & $0.375 \scriptscriptstyle \pm \scriptstyle 1.2e\text{-}3$ & $0.721 \scriptscriptstyle \pm \scriptstyle 1.0e\text{-}3$ & $0.727 \scriptscriptstyle \pm \scriptstyle 1.0e\text{-}3$ & -- & $0.8837 \scriptscriptstyle \pm \scriptstyle 1.2e\text{-}3$ & $8.947 \scriptscriptstyle \pm \scriptstyle 8.5e\text{-}3$ & $0.969 \scriptscriptstyle \pm \scriptstyle 5.1e\text{-}4$ & $\mathbf{\textcolor{red}{0.736 \scriptscriptstyle \pm \scriptstyle 2.1e\text{-}4}}$ & $\mathbf{\textcolor{red}{0.742 \scriptscriptstyle \pm \scriptstyle 1.3e\text{-}4}}$ \\
\bottomrule
\end{tabular}

    \bigskip

    \caption{Results for ensembles with standard deviations. Color notation follows \autoref{tab:S-single-models}, "top" results are defined as in \autoref{tab:node}. Best viewed in colors.}
    \label{tab:S-ensembles}
    \small
    \begin{tabular}{lccccccccccc}
\toprule
{} & CA \textdownarrow & AD \textuparrow & HE \textuparrow & JA \textuparrow & HI \textuparrow & AL \textuparrow & EP \textuparrow & YE \textdownarrow & CO \textuparrow & YA \textdownarrow & MI \textdownarrow \\
\midrule
\multicolumn{12}{c}{Baseline Neural Networks}\\
\midrule
TabNet          & $0.488 \scriptscriptstyle \pm \scriptstyle 1.8e\text{-}3$ & $0.856 \scriptscriptstyle \pm \scriptstyle 3.4e\text{-}4$ & $0.391 \scriptscriptstyle \pm \scriptstyle 3.1e\text{-}4$ & $\mathbf{0.736 \scriptscriptstyle \pm \scriptstyle 1.3e\text{-}3}$ & $0.727 \scriptscriptstyle \pm \scriptstyle 1.3e\text{-}3$ & $0.961 \scriptscriptstyle \pm \scriptstyle 2.8e\text{-}4$ & $0.8944 \scriptscriptstyle \pm \scriptstyle 6.8e\text{-}4$ & $8.728 \scriptscriptstyle \pm \scriptstyle 8.0e\text{-}3$ & $0.966 \scriptscriptstyle \pm \scriptstyle 1.5e\text{-}3$ & $0.815 \scriptscriptstyle \pm \scriptstyle 3.4e\text{-}3$ & $0.746 \scriptscriptstyle \pm \scriptstyle 3.5e\text{-}4$ \\
SNN             & $0.478 \scriptscriptstyle \pm \scriptstyle 1.0e\text{-}3$ & $0.857 \scriptscriptstyle \pm \scriptstyle 3.1e\text{-}4$ & $0.380 \scriptscriptstyle \pm \scriptstyle 1.2e\text{-}3$ & $0.727 \scriptscriptstyle \pm \scriptstyle 8.7e\text{-}4$ & $0.729 \scriptscriptstyle \pm \scriptstyle 2.2e\text{-}3$ & $0.962 \scriptscriptstyle \pm \scriptstyle 2.8e\text{-}4$ & $0.8976 \scriptscriptstyle \pm \scriptstyle 7.5e\text{-}5$ & $8.759 \scriptscriptstyle \pm \scriptstyle 1.4e\text{-}3$ & $0.966 \scriptscriptstyle \pm \scriptstyle 4.5e\text{-}4$ & $0.754 \scriptscriptstyle \pm \scriptstyle 4.0e\text{-}4$ & $0.747 \scriptscriptstyle \pm \scriptstyle 5.2e\text{-}4$ \\
AutoInt         & $\mathbf{0.459 \scriptscriptstyle \pm \scriptstyle 3.7e\text{-}3}$ & $\mathbf{\textcolor{blue}{0.860 \scriptscriptstyle \pm \scriptstyle 2.2e\text{-}4}}$ & $0.382 \scriptscriptstyle \pm \scriptstyle 3.7e\text{-}4$ & $0.733 \scriptscriptstyle \pm \scriptstyle 7.8e\text{-}4$ & $\mathbf{\textcolor{red}{0.732 \scriptscriptstyle \pm \scriptstyle 6.6e\text{-}4}}$ & $0.959 \scriptscriptstyle \pm \scriptstyle 1.7e\text{-}4$ & $0.8966 \scriptscriptstyle \pm \scriptstyle 2.5e\text{-}4$ & $8.736 \scriptscriptstyle \pm \scriptstyle 3.0e\text{-}3$ & $0.950 \scriptscriptstyle \pm \scriptstyle 1.1e\text{-}3$ & $0.758 \scriptscriptstyle \pm \scriptstyle 1.7e\text{-}4$ & $0.747 \scriptscriptstyle \pm \scriptstyle 1.5e\text{-}4$ \\
GrowNet         & $0.468 \scriptscriptstyle \pm \scriptstyle 1.4e\text{-}3$ & $0.859 \scriptscriptstyle \pm \scriptstyle 6.3e\text{-}4$ & -- & -- & $0.730 \scriptscriptstyle \pm \scriptstyle 4.1e\text{-}4$ & -- & $0.8978 \scriptscriptstyle \pm \scriptstyle 1.5e\text{-}4$ & $\mathbf{\textcolor{red}{8.683 \scriptscriptstyle \pm \scriptstyle 6.6e\text{-}3}}$ & -- & $0.756 \scriptscriptstyle \pm \scriptstyle 4.7e\text{-}4$ & $0.747 \scriptscriptstyle \pm \scriptstyle 1.4e\text{-}4$ \\
MLP             & $0.487 \scriptscriptstyle \pm \scriptstyle 7.9e\text{-}4$ & $0.855 \scriptscriptstyle \pm \scriptstyle 4.8e\text{-}4$ & $0.390 \scriptscriptstyle \pm \scriptstyle 1.4e\text{-}3$ & $0.725 \scriptscriptstyle \pm \scriptstyle 2.1e\text{-}4$ & $0.725 \scriptscriptstyle \pm \scriptstyle 3.1e\text{-}4$ & $0.960 \scriptscriptstyle \pm \scriptstyle 3.2e\text{-}4$ & $\mathbf{0.8979 \scriptscriptstyle \pm \scriptstyle 1.1e\text{-}4}$ & $8.712 \scriptscriptstyle \pm \scriptstyle 6.3e\text{-}3$ & $0.966 \scriptscriptstyle \pm \scriptstyle 9.1e\text{-}5$ & $0.753 \scriptscriptstyle \pm \scriptstyle 1.5e\text{-}4$ & $0.746 \scriptscriptstyle \pm \scriptstyle 1.4e\text{-}4$ \\
DCN2            & $0.477 \scriptscriptstyle \pm \scriptstyle 3.7e\text{-}4$ & $0.857 \scriptscriptstyle \pm \scriptstyle 3.2e\text{-}4$ & $0.388 \scriptscriptstyle \pm \scriptstyle 1.5e\text{-}3$ & $0.719 \scriptscriptstyle \pm \scriptstyle 1.5e\text{-}3$ & $0.725 \scriptscriptstyle \pm \scriptstyle 1.0e\text{-}3$ & $0.960 \scriptscriptstyle \pm \scriptstyle 4.1e\text{-}4$ & $0.8977 \scriptscriptstyle \pm \scriptstyle 4.8e\text{-}5$ & $8.800 \scriptscriptstyle \pm \scriptstyle 9.9e\text{-}3$ & $\mathbf{0.969 \scriptscriptstyle \pm \scriptstyle 6.4e\text{-}4}$ & $0.752 \scriptscriptstyle \pm \scriptstyle 7.7e\text{-}4$ & $0.746 \scriptscriptstyle \pm \scriptstyle 3.7e\text{-}4$ \\
NODE            & $0.461 \scriptscriptstyle \pm \scriptstyle 6.9e\text{-}4$ & $0.860 \scriptscriptstyle \pm \scriptstyle 7.0e\text{-}4$ & $0.361 \scriptscriptstyle \pm \scriptstyle 7.9e\text{-}4$ & $0.730 \scriptscriptstyle \pm \scriptstyle 8.4e\text{-}4$ & $0.727 \scriptscriptstyle \pm \scriptstyle 9.1e\text{-}4$ & $0.921 \scriptscriptstyle \pm \scriptstyle 1.6e\text{-}3$ & $0.8970 \scriptscriptstyle \pm \scriptstyle 3.7e\text{-}4$ & $8.716 \scriptscriptstyle \pm \scriptstyle 3.1e\text{-}3$ & $0.965 \scriptscriptstyle \pm \scriptstyle 5.0e\text{-}4$ & $\mathbf{0.750 \scriptscriptstyle \pm \scriptstyle 2.1e\text{-}5}$ & $\mathbf{0.744 \scriptscriptstyle \pm \scriptstyle 8.2e\text{-}5}$ \\
ResNet          & $0.478 \scriptscriptstyle \pm \scriptstyle 7.9e\text{-}4$ & $0.857 \scriptscriptstyle \pm \scriptstyle 4.3e\text{-}4$ & $\mathbf{0.398 \scriptscriptstyle \pm \scriptstyle 7.2e\text{-}4}$ & $0.734 \scriptscriptstyle \pm \scriptstyle 1.3e\text{-}3$ & $0.731 \scriptscriptstyle \pm \scriptstyle 8.5e\text{-}4$ & $\mathbf{0.966 \scriptscriptstyle \pm \scriptstyle 4.9e\text{-}4}$ & $0.8976 \scriptscriptstyle \pm \scriptstyle 2.7e\text{-}4$ & $8.770 \scriptscriptstyle \pm \scriptstyle 8.0e\text{-}3$ & $0.967 \scriptscriptstyle \pm \scriptstyle 6.7e\text{-}4$ & $0.751 \scriptscriptstyle \pm \scriptstyle 7.5e\text{-}5$ & $0.745 \scriptscriptstyle \pm \scriptstyle 1.9e\text{-}4$ \\
\midrule
\multicolumn{12}{c}{\architecture}\\
\midrule
FT-Transformer\textsubscript{d} & $0.454 \scriptscriptstyle \pm \scriptstyle 1.1e\text{-}3$ & $0.860 \scriptscriptstyle \pm \scriptstyle 4.9e\text{-}4$ & $0.395 \scriptscriptstyle \pm \scriptstyle 9.4e\text{-}4$ & $0.734 \scriptscriptstyle \pm \scriptstyle 7.5e\text{-}4$ & $0.731 \scriptscriptstyle \pm \scriptstyle 8.0e\text{-}4$ & $0.966 \scriptscriptstyle \pm \scriptstyle 3.9e\text{-}4$ & $0.8969 \scriptscriptstyle \pm \scriptstyle 1.9e\text{-}4$ & $8.727 \scriptscriptstyle \pm \scriptstyle 1.6e\text{-}2$ & $0.973 \scriptscriptstyle \pm \scriptstyle 3.2e\text{-}4$ & $\mathbf{\textcolor{blue}{0.747 \scriptscriptstyle \pm \scriptstyle 3.8e\text{-}4}}$ & $\mathbf{\textcolor{blue}{0.742 \scriptscriptstyle \pm \scriptstyle 3.3e\text{-}4}}$ \\
FT-Transformer  & $\mathbf{\textcolor{blue}{0.448 \scriptscriptstyle \pm \scriptstyle 7.5e\text{-}4}}$ & $0.860 \scriptscriptstyle \pm \scriptstyle 3.9e\text{-}4$ & $\mathbf{\textcolor{red}{0.398 \scriptscriptstyle \pm \scriptstyle 4.3e\text{-}4}}$ & $\mathbf{\textcolor{red}{0.739 \scriptscriptstyle \pm \scriptstyle 5.9e\text{-}4}}$ & $0.731 \scriptscriptstyle \pm \scriptstyle 7.7e\text{-}4$ & $\mathbf{\textcolor{red}{0.967 \scriptscriptstyle \pm \scriptstyle 4.8e\text{-}4}}$ & $\mathbf{\textcolor{red}{0.8984 \scriptscriptstyle \pm \scriptstyle 1.6e\text{-}4}}$ & $8.751 \scriptscriptstyle \pm \scriptstyle 9.4e\text{-}3$ & $\mathbf{\textcolor{red}{0.973 \scriptscriptstyle \pm \scriptstyle 1.1e\text{-}4}}$ & $0.747 \scriptscriptstyle \pm \scriptstyle 3.8e\text{-}4$ & $0.743 \scriptscriptstyle \pm \scriptstyle 1.1e\text{-}4$ \\
\midrule
\multicolumn{12}{c}{GBDT}\\
\midrule
CatBoost\textsubscript{d} & $0.428 \scriptscriptstyle \pm \scriptstyle 4.5e\text{-}5$ & $0.873 \scriptscriptstyle \pm \scriptstyle 4.2e\text{-}4$ & $0.386 \scriptscriptstyle \pm \scriptstyle 1.0e\text{-}3$ & $0.724 \scriptscriptstyle \pm \scriptstyle 4.8e\text{-}4$ & $0.728 \scriptscriptstyle \pm \scriptstyle 7.4e\text{-}4$ & $0.948 \scriptscriptstyle \pm \scriptstyle 9.2e\text{-}4$ & $0.8893 \scriptscriptstyle \pm \scriptstyle 2.7e\text{-}4$ & $8.885 \scriptscriptstyle \pm \scriptstyle 1.9e\text{-}3$ & $0.910 \scriptscriptstyle \pm \scriptstyle 3.0e\text{-}4$ & $0.749 \scriptscriptstyle \pm \scriptstyle 1.1e\text{-}4$ & $0.744 \scriptscriptstyle \pm \scriptstyle 4.4e\text{-}5$ \\
CatBoost        & $\mathbf{\textcolor{red}{0.423 \scriptscriptstyle \pm \scriptstyle 8.9e\text{-}4}}$ & $\mathbf{\textcolor{red}{0.874 \scriptscriptstyle \pm \scriptstyle 4.5e\text{-}4}}$ & $0.388 \scriptscriptstyle \pm \scriptstyle 2.7e\text{-}4$ & $0.727 \scriptscriptstyle \pm \scriptstyle 6.4e\text{-}4$ & $0.729 \scriptscriptstyle \pm \scriptstyle 1.6e\text{-}3$ & -- & $0.8898 \scriptscriptstyle \pm \scriptstyle 7.7e\text{-}5$ & $8.837 \scriptscriptstyle \pm \scriptstyle 3.2e\text{-}3$ & $0.968 \scriptscriptstyle \pm \scriptstyle 2.2e\text{-}5$ & $0.740 \scriptscriptstyle \pm \scriptstyle 1.7e\text{-}4$ & $\mathbf{\textcolor{red}{0.741 \scriptscriptstyle \pm \scriptstyle 7.3e\text{-}5}}$ \\
XGBoost\textsubscript{d} & $0.462 \scriptscriptstyle \pm \scriptstyle 0.0$ & $0.874 \scriptscriptstyle \pm \scriptstyle 0.0$ & $0.348 \scriptscriptstyle \pm \scriptstyle 0.0$ & $0.711 \scriptscriptstyle \pm \scriptstyle 0.0$ & $0.717 \scriptscriptstyle \pm \scriptstyle 0.0$ & $0.924 \scriptscriptstyle \pm \scriptstyle 0.0$ & $0.8799 \scriptscriptstyle \pm \scriptstyle 0.0$ & $9.192 \scriptscriptstyle \pm \scriptstyle 0.0$ & $0.964 \scriptscriptstyle \pm \scriptstyle 0.0$ & $0.761 \scriptscriptstyle \pm \scriptstyle 0.0$ & $0.751 \scriptscriptstyle \pm \scriptstyle 0.0$ \\
XGBoost         & $0.431 \scriptscriptstyle \pm \scriptstyle 3.6e\text{-}4$ & $0.872 \scriptscriptstyle \pm \scriptstyle 2.3e\text{-}4$ & $0.377 \scriptscriptstyle \pm \scriptstyle 7.6e\text{-}4$ & $0.724 \scriptscriptstyle \pm \scriptstyle 3.4e\text{-}4$ & $0.728 \scriptscriptstyle \pm \scriptstyle 5.3e\text{-}4$ & -- & $0.8861 \scriptscriptstyle \pm \scriptstyle 1.6e\text{-}4$ & $8.819 \scriptscriptstyle \pm \scriptstyle 4.0e\text{-}3$ & $0.969 \scriptscriptstyle \pm \scriptstyle 1.9e\text{-}4$ & $\mathbf{\textcolor{red}{0.732 \scriptscriptstyle \pm \scriptstyle 5.4e\text{-}5}}$ & $0.742 \scriptscriptstyle \pm \scriptstyle 1.8e\text{-}5$ \\
\bottomrule
\end{tabular}
\end{sidewaystable}

\newpage
\subsection{Additional results}

\subsubsection{Training times}
\begin{table}[h]
\setlength\tabcolsep{3.5pt}
\caption{Training times in seconds averaged over 15 runs.}
\label{tab:S-training-times}
\centering
\vspace{1em}
\begin{tabular}{lccccccccccc}
\toprule
{} & CA & AD & HE & JA & HI & AL & EP & YE & CO & YA & MI \\
\midrule
ResNet & 72 & 144 & 363 & 163 & 91 & 933 & 704 & 777 & 4026 & 923 & 1243 \\
FT-Transformer & 187 & 128 & 536 & 576 & 257 & 2864 & 934 & 1776 & 5050 & 12712 & 2857 \\
\midrule
Overhead & 2.6x & 0.9x & 1.5x & 3.5x & 2.8x & 3.1x & 1.3x & 2.3x & 1.3x & 13.8x & 2.3x \\
\bottomrule
\end{tabular}
\end{table}
For most experiments, training times can be found in the source code.
In \autoref{tab:S-training-times}, we provide the comparison between ResNet and \architecture\ in order to ``visualize'' the overhead introduced by \architecture\ compared to the main ``conventional'' DL baseline.
The big difference on the Yahoo dataset is expected because of the large number of features (700).

\subsubsection{How tuning time budget affects performance?}
In this section, we aim to answer the following questions:
\begin{itemize}[nosep]
    \item how does the relative performance of tuned models depends on tuning \textit{time} budget?
    \item does the number of tuning iterations used in the main text allow models to reach most of their potential?
\end{itemize}
The first question is important for two main reasons.
First, we have to make sure that longer tuning \textit{times} of FT-Transformer (the number of tuning \textit{iterations} is the same as for all other models) is not the reason of its strong performance.
Second, we want to test FT-Transformer in the regime of low tuning time budget.

We consider four algorithms: XGBoost (as a fast GBDT implementation), MLP (as the fastest and simplest DL model), ResNet (as a stronger but slower DL model), FT-Transformer (as the strongest and the slowest DL model).
We consider three datasets: California Housing, Adult, Higgs Small.
On each dataset, for each algorithm, we run five independent (five random seeds) hyperparameter optimizations.
Each run is constrained only by \textit{time}.
For each of the considered time budgets (15 minutes, 30 minutes, 1 hour, 2 hours, 3 hours, 4 hours, 5 hours, 6 hours), we pick the best model identified by Optuna on the validation set using no more than this time budget. Then, we report its performance and the number of Optuna iterations averaged over the five random seeds.
The results are reported in \autoref{tab:S-tuning-time-budget}.
The takeaways are as follows:
\begin{itemize}[nosep]
    \item interestingly, FT-Transformer achieves good metrics just after several randomly sampled configurations (Optuna performs simple random sampling during the first 10 (default) iterations). 
    \item FT-Transformer is slower to train, which is expected
    \item extended tuning (in terms of iterations) for other algorithms does not lead to any meaningful improvements
\end{itemize}

\begin{table}[h]
\setlength\tabcolsep{2.5pt}
\caption{Performance of tuned models with different tuning time budgets. Tuned model performance and the number of Optuna iterations (in parentheses) are reported (both metrics are averaged over five random seeds). Best results among DL models are in bold, overall best results are in bold red.}
\label{tab:S-tuning-time-budget}
\centering
\vspace{1em}
\makebox[\textwidth][c]{
    \begin{tabular}{lcccccccc}
\toprule
{} & 0.25h & 0.5h & 1h& 2h & 3h & 4h & 5h & 6h \\

\midrule
\multicolumn{9}{c}{California Housing} \\
\midrule

XGBoost & $\textbf{\textcolor{red}{0.437 (31)}}$ & $\textbf{\textcolor{red}{0.436 (56)}}$ & $\textbf{\textcolor{red}{0.434 (120)}}$ & $\textbf{\textcolor{red}{0.433 (252)}}$ & $\textbf{\textcolor{red}{0.433 (410)}}$ & $\textbf{\textcolor{red}{0.432 (557)}}$ & $\textbf{\textcolor{red}{0.433 (719)}}$ & $\textbf{\textcolor{red}{0.432 (867)}}$  \\
MLP & $0.503 (16)$ & $0.496 (42)$ & $0.493 (103)$ & $0.488 (230)$ & $0.489 (349)$ & $0.489 (466)$ & $0.488 (596)$ & $0.488 (724)$  \\
ResNet & $0.488 (7)$ & $0.487 (15)$ & $0.483 (30)$ & $0.481 (64)$ & $0.482 (101)$ & $0.482 (131)$ & $0.482 (164)$ & $0.484 (197)$  \\
FT-Transformer & $\textbf{0.466 (4)}$ & $\textbf{0.464 (9)}$ & $\textbf{0.465 (20)}$ & $\textbf{0.460 (47)}$ & $\textbf{0.458 (74)}$ & $\textbf{0.458 (99)}$ & $\textbf{0.457 (124)}$ & $\textbf{0.459 (153)}$  \\

\midrule
\multicolumn{9}{c}{Adult} \\
\midrule

XGBoost & $\textbf{\textcolor{red}{0.871 (165)}}$ & $\textbf{\textcolor{red}{0.873 (311)}}$ & $\textbf{\textcolor{red}{0.872 (638)}}$ & $\textbf{\textcolor{red}{0.872 (1296)}}$ & $\textbf{\textcolor{red}{0.872 (1927)}}$ & $\textbf{\textcolor{red}{0.872 (2478)}}$ & $\textbf{\textcolor{red}{0.872 (2999)}}$ & $\textbf{\textcolor{red}{0.872 (3500)}}$ \\
MLP & $0.856 (20)$ & $0.857 (37)$ & $0.858 (71)$ & $0.857 (130)$ & $0.856 (190)$ & $0.856 (247)$ & $0.856 (310)$ & $0.856 (375)$  \\
ResNet & $0.856 (8)$ & $0.854 (16)$ & $0.854 (32)$ & $0.856 (69)$ & $0.855 (105)$ & $0.855 (140)$ & $0.856 (174)$ & $0.855 (208)$  \\
FT-Transformer & $\textbf{0.861 (6)}$ & $\textbf{0.860 (12)}$ & $\textbf{0.859 (27)}$ & $\textbf{0.859 (52)}$ & $\textbf{0.860 (78)}$ & $\textbf{0.860 (99)}$ & $\textbf{0.860 (125)}$ & $\textbf{0.860 (148)}$  \\

\midrule
\multicolumn{9}{c}{Higgs Small} \\
\midrule

XGBoost & $0.725 (88)$ & $0.725 (153)$ & $0.724 (291)$ & $0.725 (573)$ & $0.725 (823)$ & $0.726 (1069)$ & $0.725 (1318)$ & $0.725 (1559)$ \\
MLP & $0.721 (16)$ & $0.720 (29)$ & $0.723 (62)$ & $0.722 (137)$ & $0.724 (220)$ & $0.723 (300)$ & $0.724 (375)$ & $0.724 (447)$  \\
ResNet & $0.724 (8)$ & $0.727 (14)$ & $0.727 (32)$ & $0.728 (61)$ & $0.728 (84)$ & $0.728 (107)$ & $0.728 (132)$ & $0.728 (154)$  \\
FT-Transformer & $\textbf{\textcolor{red}{0.727 (2)}}$ & $\textbf{\textcolor{red}{0.729 (5)}}$ & $\textbf{\textcolor{red}{0.728 (12)}}$ & $\textbf{\textcolor{red}{0.728 (23)}}$ & $\textbf{\textcolor{red}{0.729 (34)}}$ & $\textbf{\textcolor{red}{0.729 (44)}}$ & $\textbf{\textcolor{red}{0.730 (56)}}$ & $\textbf{\textcolor{red}{0.729 (66)}}$   \\

\bottomrule
\end{tabular}
}
\end{table}

\newpage
\subsection{\architecture}
In this section, we formally describe the details of \architecture\, its tuning and evaluation.
Also, we share additional technical experience and observations that were not used for final results in the paper but may be of interest to researchers and practitioners.

\subsubsection{Architecture}
\label{sec:S-architecture}

\textbf{Formal definition.}
\begin{equation*}
    \texttt{\architecture}(x) = \texttt{Prediction}(\texttt{Block}(\ldots(\texttt{Block}(\texttt{AppendCLS}(\texttt{FeatureTokenizer}(x))))))
\end{equation*}
\begin{align*}
    \texttt{Block}(x) & = \texttt{ResidualPreNorm}(\texttt{FFN},\ \texttt{ResidualPreNorm}(\texttt{MHSA},\ x))
    \\ \texttt{ResidualPreNorm}(\texttt{Module},\ x) & = x + \texttt{Dropout}(\texttt{Module}(\texttt{Norm}(x)))
    \\ \texttt{FFN}(x) & = \texttt{Linear}(\texttt{Dropout}(\texttt{Activation}(\texttt{Linear}(x))))
\end{align*}
We use \texttt{LayerNorm} \citep{layernorm} as the normalization.
See the main text for the description of \texttt{Prediction} and \texttt{FeatureTokenizer}.
For MHSA, we set $n_{heads} = 8$ and do not tune this parameter.

\textbf{Activation.}
Throughout the whole paper we used the ReGLU activation, since it is reported to be superior to the usually used GELU activation \citep{glu-variants,transformer-modifications}.
However, we did not observe strong difference between ReGLU and ReLU in preliminary experiments.

\textbf{Dropout rates.}
We observed that the attention dropout is always beneficial and FFN-dropout is also usually set by the tuning process to some non-zero value.
As for the final dropout of each residual branch, it is rarely set to non-zero values by the tuning process.

\textbf{PreNorm vs PostNorm.} We use the PreNorm variant of Transformer, i.e. normalizations are placed at the beginning of each residual branch.
The PreNorm variant is known for better optimization properties as opposed to the original Transformer, which is a PostNorm-Transformer \citep{prenorm,admin,tears}.
The latter one may produce better models in terms of target metrics \citep{admin}, but it usually requires additional modifications to the model and/or the training process, such as learning rate warmup or complex initialization schemes \citep{tfixup,admin}.
While the PostNorm variant can be an option for practitioners seeking for the best possible model, we use the PreNorm variant in order to keep the optimization simple and same for all models.
Note that in the PostNorm formulation the \texttt{LayerNorm} in the "Prediction" equation (see the section ``\architecture'' in the main text) should be omitted.

\subsubsection{The default configuration(s)}
\autoref{tab:S-default-config} describes the configuration of \architecture\ referred to as ``default'' in the main text.
Note that it includes hyperparameters for both the model and the optimization.
In fact, the configuration is a result of an ``educated guess'' and we did not invest much resources in its tuning.

\begin{table}[h]
    \caption{Default \architecture\ used in the main text.}
    \label{tab:S-default-config}
    \centering
    \vspace{1em}
    \begin{tabular}{c|c|l}
        Layer count &  3 & \\
        Feature embedding size & 192 & \\
        Head count & 8 & \\
        Activation \& FFN size factor & (\texttt{ReGLU}, $\nicefrac{4}{3}$) & \\
        Attention dropout & 0.2 & \\
        FFN dropout & 0.1 & \\
        Residual dropout & 0.0 & \\
        Initialization & \texttt{Kaiming} & \citep{kaiming-init} \\
        \midrule
        Parameter count & 929K & The value is given for 100 numerical features \\
        \midrule
        Optimizer & \texttt{AdamW} & \\
        Learning rate & \num{1e-4} & \\
        Weight decay & \num{1e-5} & 0.0 for Feature Tokenizer, \texttt{LayerNorm} and biases \\
    \end{tabular}
\end{table}
where ``FFN size factor'' is a ratio of the \texttt{FFN}'s hidden size to the feature embedding size.

We also designed a heuristic scaling rule to produce ``default'' configurations with the number of layers from one to six.
We applied it on the Epsilon and Yahoo datasets in order to reduce the number of tuning iterations.
However, we did not dig into the topic and our scaling rule may be suboptimal, see \citet{which-transformer-fits-my-data} for a theoretically sound scaling rule.

\tuningparagraph{tab:S-transformer-space}\
For Epsilon, however, we iterated over several ``default'' configurations using a heuristic scaling rule, since the full tuning procedure turned out to be too time consuming.
For Yahoo, we did not perform tuning at all, since the default configuration already performed well.
In the main text, for FT-Transformer on Yahoo, we report the result of the default FT-Transformer.

\begin{table}[h]
\centering
\caption{\architecture\ hyperparameter space. Here (A) = \{CA, AD, HE, JA, HI\} and \\ (B) = \{AL, YE, CO, MI\} }
\label{tab:S-transformer-space}
\vspace{1em}
{\renewcommand{\arraystretch}{1.2}
\begin{tabular}{ll}
    \toprule
    Parameter & (Datasets) Distribution \\
    \midrule
    \# Layers & (A) $\mathrm{UniformInt}[1,4]$, (B) $\mathrm{UniformInt}[1,6]$ \\
    Feature embedding size & (A,B) $\mathrm{UniformInt}[64,512]$ \\
    Residual dropout &  (A) $\{0, \mathrm{Uniform}[0, 0.2] \}$, (B) $\mathrm{Const}(0.0)$ \\
    Attention dropout &  (A,B) $\mathrm{Uniform}[0, 0.5]$ \\
    FFN dropout & (A,B) $\mathrm{Uniform}[0, 0.5]$ \\
    FFN factor & (A) $\mathrm{Uniform}[\nicefrac{2}{3}, \nicefrac{8}{3}]$, (B) $\mathrm{Const}(\nicefrac{4}{3})$ \\
    Learning rate & (A) $\mathrm{LogUniform}[1e\text{-}5, 1e\text{-}3]$, (B) $\mathrm{LogUniform}[3e\text{-}5, 3e\text{-}4]$ \\
    Weight decay & (A,B) $\mathrm{LogUniform}[1e\text{-}6, 1e\text{-}3]$ \\
    \midrule
    \# Iterations & (A) 100, (B) 50 \\
    \bottomrule
\end{tabular}}
\end{table}

\subsubsection{Training}
On the Epsilon dataset, we scale \architecture\ using the technique proposed by \citet{linformer} with the ``headwise'' sharing policy; we set the projection dimension to 128.
We follow the popular ``transformers'' library \citep{hf-transformers} and do not apply weight decay to \tokenizer, biases in linear layers and normalization layers.

\subsection{Models}
In this section, we describe the implementation details for all models. See \autoref{sec:S-architecture} for details on \architecture.

\subsubsection{ResNet}

\textbf{Architecture.}
The architecture is formally described in the main text.

We tested several configurations and observed measurable difference in performance between all of them. We found the ones with ``clear main path'' (i.e. with all normalizations (except the last one) placed only in residual branches as in \citet{preactivation} or \citet{prenorm}) to perform better. As expected, it is also easier for them to train deeper configurations. We found the block design inspired by Transformer \citep{transformer} to perform better or on par with the one inspired by the ResNet from computer vision \citep{resnet}.

We observed that in the ``optimal'' configurations (the result of the hyperparameter optimization process) the inner dropout rate (not the last one) of one block was usually set to higher values compared to the outer dropout rate. Moreover, the latter one was set to zero in many cases.

\textbf{Implementation.} Ours, see the source code.

\tuningparagraph{tab:S-resnet-space}

\begin{table}[h]
\centering
\caption{ResNet hyperparameter space. Here (A) = \{CA, AD, HE, JA, HI, AL\} and \\ (B) = \{EP, YE, CO, YA, MI\} }
\label{tab:S-resnet-space}
\vspace{1em}
{\renewcommand{\arraystretch}{1.2}
\begin{tabular}{ll}
    \toprule
    Parameter & (Datasets) Distribution \\
    \midrule
    \# Layers & (A) $\mathrm{UniformInt}[1,8]$, (B) $\mathrm{UniformInt}[1,16]$ \\
    Layer size & (A) $\mathrm{UniformInt}[64,512]$, (B) $\mathrm{UniformInt}[64,1024]$ \\
    Hidden factor & (A,B) $\mathrm{Uniform}[1, 4]$\\
    Hidden dropout &  (A,B) $\mathrm{Uniform}[0, 0.5]$ \\
    Residual dropout & (A,B) $\{0, \mathrm{Uniform}[0, 0.5]\}$ \\
    Learning rate & (A,B) $\mathrm{LogUniform}[1e\text{-}5, 1e\text{-}2]$ \\
    Weight decay & (A,B) $\{0, \mathrm{LogUniform}[1e\text{-}6, 1e\text{-}3] \}$ \\
    Category embedding size & (\{AD\}) $\mathrm{UniformInt}[64, 512]$ \\
    \midrule
    \# Iterations & 100 \\
    \bottomrule
\end{tabular}}
\end{table}

\subsubsection{MLP}

\textbf{Architecture.}
The architecture is formally described in the main text.

\textbf{Implementation.}
Ours, see the source code.

\tuningparagraph{tab:S-mlp-space}\ Note that the size of the first and the last layers are tuned and set separately, while the size for ``in-between'' layers is the same for all of them.

\begin{table}[h]
\centering
\caption{MLP hyperparameter space. Here (A) = \{CA, AD, HE, JA, HI, AL\} and \\ (B) = \{EP, YE, CO, YA, MI\} }
\label{tab:S-mlp-space}
\vspace{1em}
{\renewcommand{\arraystretch}{1.2}
\begin{tabular}{ll}
    \toprule
    Parameter & (Datasets) Distribution \\
    \midrule
    \# Layers & (A) $\mathrm{UniformInt}[1,8]$, (B) $\mathrm{UniformInt}[1,16]$ \\
    Layer size & (A) $\mathrm{UniformInt}[1,512]$, (B) $\mathrm{UniformInt}[1,1024]$ \\
    Dropout &  (A,B) $\{0, \mathrm{Uniform}[0, 0.5]\}$ \\
    Learning rate & (A,B) $\mathrm{LogUniform}[1e\text{-}5, 1e\text{-}2]$ \\
    Weight decay & (A,B) $\{0, \mathrm{LogUniform}[1e\text{-}6, 1e\text{-}3] \}$ \\
    Category embedding size & (\{AD\}) $\mathrm{UniformInt}[64, 512]$ \\
    \midrule
    \# Iterations & 100 \\
    \bottomrule
\end{tabular}}
\end{table}

\subsubsection{XGBoost}

\textbf{Implementation.} We fix and do not tune the following hyperparameters:
\begin{itemize}
    \itemsep0em
    \item $\texttt{booster} = \text{"gbtree"}$
    \item $\texttt{early-stopping-rounds} = 50$
    \item $\texttt{n-estimators} = 2000$
\end{itemize}

\tuningparagraph{tab:S-xgboost-space}

\begin{table}[h]
\centering
\caption{XGBoost hyperparameter space. Here (A) = \{CA, AD, HE, JA, HI\} and \\ (B) = \{EP, YE, CO, YA, MI\} }
\label{tab:S-xgboost-space}
\vspace{1em}
{\renewcommand{\arraystretch}{1.2}
\begin{tabular}{ll}
    \toprule
    Parameter & (Datasets) Distribution \\
    \midrule
    Max depth & (A) $\mathrm{UniformInt[3,10]}$, (B) $\mathrm{UniformInt[6,10]}$ \\
    Min child weight & (A,B) $\mathrm{LogUniform}[1e\text{-}8, 1e5]$ \\
    Subsample & (A,B) $\mathrm{Uniform}[0.5, 1]$ \\
    Learning rate & (A,B) $\mathrm{LogUniform}[1e\text{-}5, 1]$ \\
    Col sample by level & (A,B) $\mathrm{Uniform}[0.5, 1]$ \\
    Col sample by tree & (A,B) $\mathrm{Uniform}[0.5, 1]$ \\ 
    Gamma & (A,B) $\{0, \mathrm{LogUniform}[1e\text{-}8, 1e2]\}$ \\ 
    Lambda & (A,B) $\{0, \mathrm{LogUniform}[1e\text{-}8, 1e2]\}$ \\ 
    Alpha & (A,B) $\{0, \mathrm{LogUniform}[1e\text{-}8, 1e2]\}$ \\
    \midrule
    \# Iterations & 100 \\
    \bottomrule
\end{tabular}}
\end{table}

\subsubsection{CatBoost}

\textbf{Implementation.} We fix and do not tune the following hyperparameters:
\begin{itemize}
    \itemsep0em
    \item $\texttt{early-stopping-rounds} = 50$
    \item $\texttt{od-pval} = 0.001$
    \item $\texttt{iterations} = 2000$
\end{itemize}

\tuningparagraph{tab:S-catboost-space}\
We set the \texttt{task\_type} parameter to ``GPU'' (the tuning was unacceptably slow on CPU).

\begin{table}[h]
\centering
\caption{CatBoost hyperparameter space. Here (A) = \{CA, AD, HE, JA, HI\} and \\ (B) = \{EP, YE, CO, YA, MI\} }
\label{tab:S-catboost-space}
\vspace{1em}
{\renewcommand{\arraystretch}{1.2}
\begin{tabular}{ll}
    \toprule
    Parameter & (Datasets) Distribution \\
    \midrule
    Max depth & (A) $\mathrm{UniformInt[3,10]}$, (B) $\mathrm{UniformInt[6,10]}$ \\
    Learning rate & (A,B) $\mathrm{LogUniform}[1e\text{-}5, 1]$ \\
    Bagging temperature & (A,B) $\mathrm{Uniform}[0, 1]$ \\
    L2 leaf reg  & (A,B) $\mathrm{LogUniform}[1, 10]$ \\
    Leaf estimation iterations & (A,B) $\mathrm{UniformInt}[1, 10]$ \\
    \midrule
    \# Iterations & 100 \\
    \bottomrule
\end{tabular}}
\end{table}

\textbf{Evaluation.} We set the \texttt{task\_type} parameter to ``CPU'', since for the used version of the CatBoost library it is crucial for performance in terms of target metrics.

\subsubsection{SNN}

\textbf{Implementation.} Ours, see the source code.

\tuningparagraph{tab:S-snn-space}

\begin{table}[h]
\centering
\caption{SNN hyperparameter space. Here (A) = \{CA, AD, HE, JA, HI, AL\} and \\ (B) = \{EP, YE, CO, YA, MI\} }
\label{tab:S-snn-space}
\vspace{1em}
{\renewcommand{\arraystretch}{1.2}
\begin{tabular}{ll}
    \toprule
    Parameter & (Datasets) Distribution \\
    \midrule
    \# Layers & (A) $\mathrm{UniformInt}[2,16]$, (B) $\mathrm{UniformInt}[2,32]$ \\
    Layer size & (A) $\mathrm{UniformInt}[1,512]$, (B) $\mathrm{UniformInt}[1,1024]$ \\
    Dropout &  (A,B) $\{0, \mathrm{Uniform}[0, 0.1]\}$ \\
    Learning rate & (A,B) $\mathrm{LogUniform}[1e\text{-}5, 1e\text{-}2]$ \\
    Weight decay & (A,B) $\{0, \mathrm{LogUniform}[1e\text{-}5, 1e\text{-}3] \}$ \\
    Category embedding size & (\{AD\}) $\mathrm{UniformInt}[64, 512]$ \\
    \midrule
    \# Iterations & 100 \\
    \bottomrule
\end{tabular}}
\end{table}

\subsubsection{NODE}

\textbf{Implementation.}
We used the official implementation: \url{https://github.com/Qwicen/node}.

\textbf{Tuning.}
We iterated over the parameter grid from the original paper \citep{node} plus the default configuration from the original paper.
For multiclass datasets, we set the tree dimension being equal to the number of classes.
For the Helena and ALOI datasets there was no tuning since NODE does not scale to classification problems with a large number of classes (for example, the minimal non-default configuration of NODE contains 600M+ parameters on the Helena dataset), so the reported results for these datasets are obtained with the default configuration.

\subsubsection{TabNet}

\textbf{Implementation.} We used the official implementation: \\\url{https://github.com/google-research/google-research/tree/master/tabnet}.
\\We always set $\texttt{feature-dim}$ equal to $\texttt{output-dim}$.
We also fix and do not tune the following hyperparameters (let A = \{CA, AD\}, B = \{HE, JA, HI, AL\}, C = \{EP, YE, CO, YA, MI\}):
\begin{itemize}
    \itemsep0em
    \item $\texttt{virtual-batch-size} = \mathrm{(A)}\ 2048, \mathrm{(B)}\ 8192, \mathrm{(C)}\ 16384$
    \item $\texttt{batch-size} = \mathrm{(A)}\ 256, \mathrm{(B)}\ 512, \mathrm{(C)}\ 1024$
\end{itemize}

\tuningparagraph{tab:S-tabnet-space}

\begin{table}[h]
\centering
\caption{TabNet hyperparameter space.}
\label{tab:S-tabnet-space}
\vspace{1em}
{\renewcommand{\arraystretch}{1.2}
\begin{tabular}{ll}
    \toprule
    Parameter & Distribution \\
    \midrule
    \# Decision steps & $\mathrm{UniformInt}[3,10]$ \\
    Layer size & $\{8, 16, 32, 64, 128\}$ \\
    Relaxation factor & $\mathrm{Uniform}[1,2]$ \\
    Sparsity loss weight & $\mathrm{LogUniform}[1e\text{-}6, 1e\text{-}1]$ \\
    Decay rate & $\mathrm{Uniform}[0.4, 0.95]$ \\
    Decay steps & $\{100, 500, 2000\}$ \\
    Learning rate & $\mathrm{Uniform}[1e\text{-}3, 1e\text{-}2]$ \\
    \midrule
    \# Iterations & 100 \\
    \bottomrule
\end{tabular}}
\end{table}

\subsubsection{GrowNet}

\textbf{Implementation.}
We used the official implementation: \url{https://github.com/sbadirli/GrowNet}.
Note that it does not support multiclass problems, hence the gaps in the main tables for multiclass problems.
We use no more than $40$ small MLPs, each MLP has $2$ hidden layers, boosting rate is learned -- as suggested by the authors.

\tuningparagraph{tab:S-grownet-space}

\begin{table}[h]
\centering
\caption{GrowNet hyperparameter space.}
\label{tab:S-grownet-space}
\vspace{1em}
{\renewcommand{\arraystretch}{1.2}
\begin{tabular}{ll}
    \toprule
    Parameter & (Datasets) Distribution \\
    \midrule
    Correct epochs & (all) $\{1, 2\}$ \\
    Epochs per stage & (all) $\{1, 2\}$ \\
    Hidden dimension & (all) $\mathrm{UniformInt}[32, 512]$ \\
    Learning rate & (all) $\mathrm{LogUniform}[1e\text{-}5, 1e\text{-}2]$ \\
    Weight decay & (all) $\{0, \mathrm{LogUniform}[1e\text{-}6, 1e\text{-}3] \}$ \\
    Category embedding size & (\{AD\}) $\mathrm{UniformInt}[32, 512]$ \\
    \midrule
    \# Iterations & 100 \\
    \bottomrule
\end{tabular}}
\end{table}

\subsubsection{DCN V2}

\textbf{Architecture.}
There are two variats of DCN V2, namely, ``stacked'' and ``parallel''.
We tuned and evaluated both and did not observe strong superiority of any of them.
We report numbers for the ``parallel'' variant as it was slightly better on large datasets. 

\textbf{Implementation.}
Ours, see the source code.

\tuningparagraph{tab:S-dcn2-space}

\begin{table}[h]
\centering
\caption{DCN V2 hyperparameter space. Here (A) = \{CA, AD, HE, JA, HI, AL\} and \\ (B) = \{EP, YE, CO, YA, MI\} }
\label{tab:S-dcn2-space}
\vspace{1em}
{\renewcommand{\arraystretch}{1.2}
\begin{tabular}{ll}
    \toprule
    Parameter & (Datasets) Distribution \\
    \midrule
    \# Cross layers & (A) $\mathrm{UniformInt}[1,8]$, (B) $\mathrm{UniformInt}[1,16]$ \\
    \# Hidden layers & (A) $\mathrm{UniformInt}[1,8]$, (B) $\mathrm{UniformInt}[1,16]$ \\
    Layer size & (A) $\mathrm{UniformInt}[64,512]$, (B) $\mathrm{UniformInt}[64,1024]$ \\
    Hidden dropout &  (A,B) $\mathrm{Uniform}[0, 0.5]$ \\
    Cross dropout &  (A,B) $\{0, \mathrm{Uniform}[0, 0.5]\}$ \\
    Learning rate & (A,B) $\mathrm{LogUniform}[1e\text{-}5, 1e\text{-}2]$ \\
    Weight decay & (A,B) $\{0, \mathrm{LogUniform}[1e\text{-}6, 1e\text{-}3] \}$ \\
    Category embedding size & (\{AD\}) $\mathrm{UniformInt}[64, 512]$ \\
    \midrule
    \# Iterations & 100 \\
    \bottomrule
\end{tabular}}
\end{table}

\subsubsection{AutoInt}

\textbf{Implementation.}
Ours, see the source code.
We mostly follow the original paper \citep{autoint}, however, it turns out to be necessary to introduce some modifications such as normalization in order to make the model competitive.
We fix $n_{heads} = 2$ as recommended in the original paper.

\tuningparagraph{tab:S-autoint-space}

\begin{table}[h]
\centering
\caption{AutoInt hyperparameter space. Here (A) = \{CA, AD, HE, JA, HI\} and \\ (B) = \{AL, YE, CO, MI\} }
\label{tab:S-autoint-space}
\vspace{1em}
{\renewcommand{\arraystretch}{1.2}
\begin{tabular}{ll}
    \toprule
    Parameter & (Datasets) Distribution \\
    \midrule
    \# Layers & (A,B) $\mathrm{UniformInt}[1,6]$\\
    Feature embedding size & (A,B) $\mathrm{UniformInt}[8,64]$ \\
    Residual dropout &  (A) $\{0, \mathrm{Uniform}[0.0, 0.2] \}$, (B) $\mathrm{Const}(0.0)$ \\
    Attention dropout &  (A,B) $\mathrm{Uniform}[0.0, 0.5]$ \\
    Learning rate & (A) $\mathrm{LogUniform}[1e\text{-}5, 1e\text{-}3]$, (B) $\mathrm{LogUniform}[3e\text{-}5, 3e\text{-}4]$ \\
    Weight decay & (A,B) $\mathrm{LogUniform}[1e\text{-}6, 1e\text{-}3]$ \\
    \midrule
    \# Iterations & (A) 100, (B) 50 \\
    \bottomrule
\end{tabular}}
\end{table}

\subsection{Analysis}

\subsubsection{When \architecture\ is better than ResNet?}

\textbf{Data.} Train, validation and test set sizes are $500\,000$, $50\,000$ and $100\,000$ respectively. One object is generated as $x \sim \mathcal{N}(0, I_{100})$. For each object, the first 50 features are used for target generation and the remaining 50 features play the role of ``noise''.

$f_{DL}$. The function is implemented as an MLP with three hidden layers, each of size $256$. Weights are initialized with Kaiming initialization \citep{kaiming-init}, biases are initialized with the uniform distribution $\mathcal{U}(-a,\ a)$, where $a = d_{input}^{-0.5}$. All the parameters are fixed after initialization and are not trained.

$f_{GBDT}$. The function is implemented as an average prediction of $30$ randomly constructed decision trees. The construction of one random decision tree is demonstrated in \autoref{alg:S-random-tree-construction}. The inference process for one decision tree is the same as for ordinary decision trees.

\textbf{CatBoost.} We use the default hyperparameters.

\textbf{\architecture.} We use the default hyperparameters. Parameter count: $930K$.

\textbf{ResNet.} Residual block count: $4$. Embedding size: $256$. Dropout rate inside residual blocks: $0.5$. Parameter count: $820K$.

\begin{algorithm}[h]
\normalsize{
\SetAlgoLined
\KwResult{Random Decision Tree}
set of leaves $L = \{\texttt{root}\}$\;
\texttt{depths} - mapping from nodes to their depths\;
\texttt{left} - mapping from nodes to their left children\;
\texttt{right} - mapping from nodes to their right children\;
\texttt{features} - mapping from nodes to splitting features\;
\texttt{thresholds} - mapping from nodes to splitting thresholds\;
\texttt{values} - mapping from leaves to their associated values\;
$n = 0$ - number of nodes\;
$k = 100$ - number of features\;
\While{$n < 100$}{
    randomly choose leaf $z$ from $L$ s.t. $\texttt{depths}[z] < 10$\;
    $\texttt{features}[z] \sim \texttt{UniformInt}[1,\ \ldots,\ k]$\;
    $\texttt{thresholds}[z] \sim \mathcal{N}(0,\ 1)$\;
    add two new nodes $l$ and $r$ to $L$\;
    remove $z$ from $L$\;
    unset $\texttt{values}[z]$\;
    $\texttt{left}[z] = l$\;
    $\texttt{right}[z] = r$\;
    $\texttt{depths}[l] = \texttt{depths}[r] = \texttt{depths}[z] + 1$\;
    $\texttt{values}[l] \sim \mathcal{N}(0,\ 1)$\;
    $\texttt{values}[r] \sim \mathcal{N}(0,\ 1)$\;
    $n = n + 2$\;
}
return Random Decision Tree as \{$L$, \texttt{left}, \texttt{right}, \texttt{features}, \texttt{thresholds}, \texttt{values}\}.
\caption{Construction of one random decision tree.}
\label{alg:S-random-tree-construction}
}
\end{algorithm}

\subsubsection{Ablation study}
\autoref{tab:S-ablation} is a more detailed version of the corresponding table from the main text.

\begin{table}[H]
    \centering
    \setlength\tabcolsep{2.4pt}
    \caption{The results of the comparison between \architecture\ and two attention-based alternatives. Means and standard deviations over 15 runs are reported}
    \vspace{1em}
    \label{tab:S-ablation}
    \footnotesize {
        \makebox[\textwidth][c]{
            \begin{tabular}{lcccccccc}
\toprule
{} & CA \textdownarrow & HE \textuparrow & JA \textuparrow & HI \textuparrow & AL \textuparrow & YE \textdownarrow & CO \textuparrow & MI \textdownarrow \\
\midrule
AutoInt         & $0.474 \scriptscriptstyle \pm \scriptstyle 3.3e\text{-}3$ & $0.372 \scriptscriptstyle \pm \scriptstyle 2.5e\text{-}3$ & $0.721 \scriptscriptstyle \pm \scriptstyle 2.3e\text{-}3$ & $0.725 \scriptscriptstyle \pm \scriptstyle 1.7e\text{-}3$ & $0.945 \scriptscriptstyle \pm \scriptstyle 1.3e\text{-}3$ & $8.882 \scriptscriptstyle \pm \scriptstyle 3.3e\text{-}2$ & $0.934 \scriptscriptstyle \pm \scriptstyle 3.5e\text{-}3$ & $0.750 \scriptscriptstyle \pm \scriptstyle 6.1e\text{-}4$ \\
FT-Transformer (w/o feature biases) & $0.470 \scriptscriptstyle \pm \scriptstyle 5.7e\text{-}3$ & $0.381 \scriptscriptstyle \pm \scriptstyle 1.6e\text{-}3$ & $0.724 \scriptscriptstyle \pm \scriptstyle 3.9e\text{-}3$ & $\mathbf{0.727 \scriptscriptstyle \pm \scriptstyle 1.9e\text{-}3}$ & $0.958 \scriptscriptstyle \pm \scriptstyle 1.2e\text{-}3$ & $\mathbf{8.843 \scriptscriptstyle \pm \scriptstyle 2.5e\text{-}2}$ & $0.964 \scriptscriptstyle \pm \scriptstyle 6.2e\text{-}4$ & $0.751 \scriptscriptstyle \pm \scriptstyle 5.6e\text{-}4$ \\
FT-Transformer  & $\mathbf{0.459 \scriptscriptstyle \pm \scriptstyle 3.5e\text{-}3}$ & $\mathbf{0.391 \scriptscriptstyle \pm \scriptstyle 1.2e\text{-}3}$ & $\mathbf{0.732 \scriptscriptstyle \pm \scriptstyle 2.0e\text{-}3}$ & $\mathbf{0.729 \scriptscriptstyle \pm \scriptstyle 1.5e\text{-}3}$ & $\mathbf{0.960 \scriptscriptstyle \pm \scriptstyle 1.1e\text{-}3}$ & $\mathbf{8.855 \scriptscriptstyle \pm \scriptstyle 3.1e\text{-}2}$ & $\mathbf{0.970 \scriptscriptstyle \pm \scriptstyle 6.6e\text{-}4}$ & $\mathbf{0.746 \scriptscriptstyle \pm \scriptstyle 4.9e\text{-}4}$ \\
\bottomrule
\end{tabular}
        }
    }
\end{table}

\subsection{Additional datasets}
Here, we report results for some datasets that turned out to be non-informative benchmarks, that is, where all models perform similarly. We report the average results over 15 random seeds for single models that are tuned and trained under the same protocol as described in the main text. The datasets include Bank \citep{dataset-bank}, Kick \footnote{\url{https://www.kaggle.com/c/DontGetKicked}}, MiniBooNe \footnote{\url{https://archive.ics.uci.edu/ml/datasets/MiniBooNE+particle+identification}}, Click \footnote{\url{http://www.kdd.org/kdd-cup/view/kdd-cup-2012-track-2}}. The dataset properties are given in \autoref{tab:S-additional-datasets} and the results are reported in \autoref{tab:S-additional-results}.

\begin{table}[H]
    \setlength\tabcolsep{2.4pt}
    \centering
    \caption{Additional datasets}
    \label{tab:S-additional-datasets}
    \vspace{1em}
    \begin{tabular}{lcccc}
    \toprule
    Dataset & \# objects & \# Num & \# Cat & Task type (metric) \\
    \midrule
    Bank             & 45211              & 7                       & 9                       & Binclass (accuracy)         \\
    Kick             & 72983              & 14                      & 18                      & Binclass (accuracy)         \\
    MiniBooNe        & 130064             & 50                      & 0                       & Binclass (accuracy)         \\
    Click            & 1000000            & 3                       & 8                       & Binclass (accuracy)        
    \end{tabular}
\end{table}

\begin{table}[H]
\setlength\tabcolsep{5pt}
\centering
\caption{
    Results for single models on additional datasets.
}
\label{tab:S-additional-results}
\vspace{1em}
{\footnotesize
\begin{tabular}{lcccc}
\toprule
{}                       & Bank            & Kick            & MiniBooNE       & Click           \\
\midrule
SNN                      & 0.9076 (0.0016) & 0.9014 (0.0007) & 0.9493 (0.0006) & 0.6613 (0.0006) \\
Grownet                  & 0.9093 (0.0012) & 0.9016 (0.0006) & 0.9494 (0.0007) & 0.6614 (0.0009) \\
DCNv2                    & 0.9085 (0.0010) & 0.9014 (0.0007) & 0.9496 (0.0005) & 0.6615 (0.0003) \\
AutoInt                  & 0.9065 (0.0014) & 0.9005 (0.0005) & 0.9478 (0.0008) & 0.6614 (0.0005) \\
MLP                      & 0.9059 (0.0014) & 0.9012 (0.0004) & 0.9501 (0.0006) & 0.6617 (0.0006) \\
ResNet                   & 0.9072 (0.0014) & 0.9017 (0.0005) & 0.9508 (0.0006) & 0.6612 (0.0007) \\
FT-Transformer           & 0.9090 (0.0014) & 0.9016 (0.0003) & 0.9491 (0.0007) & 0.6606 (0.0009) \\
FT-Transformer (default) & 0.9088 (0.0013) & 0.9013 (0.0006) & 0.9476 (0.0007) & 0.6610 (0.0007) \\
CatBoost                 & 0.9068 (0.0015) & 0.9021 (0.0009) & 0.9465 (0.0005) & 0.6635 (0.0002) \\
XgBoost                  & 0.9087 (0.0009) & 0.9034 (0.0003) & 0.9461 (0.0005) & 0.6399 (0.0006)
\end{tabular}
}

\end{table}

\end{document}